\documentclass{article} 
\usepackage[T1]{fontenc}
\usepackage{iclr2026_conference,times}

\usepackage{amsmath,amsfonts,bm}

\def\eqref#1{equation~\ref{#1}}

\def\1{\bm{1}}

\DeclareMathAlphabet{\mathsfit}{\encodingdefault}{\sfdefault}{m}{sl}
\SetMathAlphabet{\mathsfit}{bold}{\encodingdefault}{\sfdefault}{bx}{n}

\usepackage{booktabs}
\usepackage{url}
\usepackage{microtype}
\usepackage[table,rgb]{xcolor}
\definecolor{darkblue}{rgb}{0, 0, 0.5}
\usepackage[colorlinks=true, citecolor=darkblue, linkcolor=darkblue, urlcolor=darkblue]{hyperref}
\usepackage{amsmath}
\usepackage{bm}
\usepackage[nameinlink,capitalize,noabbrev]{cleveref}
\usepackage{graphicx}
\usepackage{inconsolata}
\usepackage{nicefrac}
\usepackage{subcaption}
\usepackage{listings}
\usepackage{enumitem}
\usepackage[framemethod=TikZ]{mdframed} 
\definecolor{light-gray}{gray}{0.97}
\newcommand*{\defeq}{\stackrel{\scalebox{0.5}{\text{def}}}{=}}

\title{Fluent Alignment with Disfluent Judges:\\
Post-training for Lower-resource Languages}

\author{David Samuel\hspace{1.5em}Lilja Øvrelid\hspace{1.5em}Erik Velldal\hspace{1.5em}Andrey Kutuzov \\
Language Technology Group, University of Oslo\\
\texttt{davisamu@ifi.uio.no}\vspace{-1em}}

\iclrfinalcopy 
\begin{document}

\maketitle

\begin{abstract}
We propose a post-training method for lower-resource languages that preserves the fluency of language models even when aligned by disfluent reward models. Preference optimization is now a well-researched topic, but previous work has mostly addressed models for English and Chinese. Lower-resource languages lack both datasets written by native speakers and instruction-tuned language models capable of generating fluent synthetic data. To address this, we focus on developing a fluent preference-aligned language model without any instruction-tuning data in the target language. Our approach uses an on-policy training method, which we compare with two common alternatives: supervised finetuning on machine-translated data and multilingual finetuning. We conduct a case study on Norwegian Bokmål and evaluate fluency through native-speaker assessments. The results show that the on-policy aspect is crucial and outperforms the alternatives without relying on any hard-to-obtain data.
\vspace{-1em}
\end{abstract}

\section{Introduction}

Instruction-tuning and preference-optimization have become a cornerstone of modern language models, enabling base models to follow instructions and engage in helpful dialogue. However, this progress has been largely confined to high-resource languages like English and Chinese, which benefit from extensive human-written datasets and sophisticated language models capable of generating fluent synthetic data. Lower-resource languages face a fundamental challenge: they lack both instruction datasets written by native speakers and fluent instruction-tuned models that could generate high-quality training data \citep{guo-etal-2025-large}. This work addresses a critical question for the democratization of language technology: how can we create fluent preference-aligned language models for lower-resource languages without any instruction-tuning dataset in the target language?\footnote{\textit{Fluency} refers to the linguistic quality of text that makes it natural, grammatical, and easy to read. It should read as though written by a native speaker. It is independent of other qualities such as factual accuracy.}

Current approaches to post-training language models for lower-resource languages mostly rely on static and predefined instruction-tuning datasets \citep{10386605, chouikhi2024gemmarenhancingllmsarabic, lim2025expandingfoundationallanguagecapabilities}, which are usually machine-translated from English \citep{pipatanakul2023typhoon, santilli-rodola-2023-camoscio, ranaldi-pucci-2023-english, ustun-etal-2024-aya, nguyen-etal-2024-seallms, bari2025allam, zosa-etal-2025-got}. While this approach shows promising results when evaluated on standard NLP benchmarks, the translation process introduces subtle linguistic artifacts -- \textit{translationese} -- that make the resulting models disfluent in the target language \citep{yu-etal-2022-translate, dutta-chowdhury-etal-2022-towards}. Translationese is produced even by professional human translators, and machine-translation models are negatively impacted by it to an even greater extent \citep{bizzoni-etal-2020-human}. Concurrent work by \cite{kunz2026preferencesidiomaticlanguageacquired} has demonstrated that even short exposure to such data leads to a rapid loss of fluency. Thus, we believe that post-training for lower-resource languages needs to shift away from such inherently flawed data.

Recent advances in reinforcement learning from AI feedback \citep[RLAIF;][]{2022constitutionalai} offer a potential solution to this challenge. In on-policy reinforcement learning, the model learns from its own generated responses rather than from fixed datasets. This means we can potentially avoid exposing it to disfluent text altogether. The key insight is that a model that has learned fluent generation through extensive pretraining on native texts can maintain this fluency as long as it is never trained on unnatural examples during the alignment phase.

In this work, we propose a fluency-aware post-training method that leverages on-policy reinforcement learning to align language models for lower-resource languages without compromising their linguistic naturalness. Crucially, we never train the model on any translated responses, preserving the fluent generation capabilities learned during pretraining. We demonstrate that even a disfluent judge model can successfully guide a fluent policy, as long as the judge understands the target language well enough to evaluate response quality.

To validate this approach, we need a language that is genuinely under-resourced for post-training yet has enough infrastructure for rigorous evaluation. \textit{Norwegian (Bokmål)}, a language with about 5 million speakers, fits this profile well: it lacks manually written instruction datasets, yet the research community has pretrained several Norwegian language models that can be leveraged for post-training. Lastly, we employed five native Norwegian speakers for accurate manual evaluation of fluency.

In the native-speaker evaluations, our on-policy approach is preferred over supervised finetuning on machine-translated data in 67.5\% of pairwise comparisons while requiring no instruction-tuning data in Norwegian. Our main contributions are:
\vspace{-0.7em}
\begin{itemize}[align=parleft,left=1em..2em]
\setlength\itemsep{0.1em}
\item We propose an on-policy reinforcement learning method for post-training in lower-resource languages that maintains fluency without requiring any instruction datasets in the target language.
\item We demonstrate through extensive human evaluation with native Norwegian speakers that on-policy training produces more fluent models than supervised finetuning on translated data.
\item We show that fluent aligned models can be bootstrapped using disfluent judges, enabling post-training for languages without existing fluent instruction-tuned models.
\item We provide comprehensive ablations revealing the critical importance of avoiding any exposure to translated responses -- even in small quantities -- during training.
\end{itemize}

\section{Fluency-aware post-training}
\label{sec:method}

This section describes our proposed method for post-training language models on a target lower-resource language without any dedicated dataset in that language. The overall approach consists of three stages: i) pretraining on the target language, ii) short SFT alignment on English, and iii) on-policy alignment on the target language.

The key principle is to \textit{never train the language model on any unnatural text}.

\paragraph{Pretraining on target language} The first stage is essential to ensure that the base language model learns all necessary linguistic knowledge and is able to generate fluent outputs in the target language. Our study focuses on the subsequent training stages and does not cover this pretraining stage -- that has already been studied in detail by \cite{gururangan-etal-2020-dont, ibrahim2024simple, kim2024efficienteffectivevocabularyexpansion, samuel-etal-2025-small}. Specifically for Norwegian, we build upon the multi-stage continual pretraining from \cite{samuel-etal-2025-small}.

\paragraph{Short SFT alignment on English} The second stage is crucial for teaching the model to respond to user prompts and follow the specific chat format (\Cref{app:chat-template}). Typically, supervised finetuning would be done on a carefully curated set of conversations in the target language, but such resources are usually not available for lower-resource languages. We therefore opt for a short alignment on a small high-quality English dataset. Specifically, we use the 1\,000 curated prompt-response pairs from LIMA \citep{10.5555/3666122.3668522} and train the base model on this dataset for one epoch. The short training ensures that the model does not catastrophically forget its innate knowledge of the target language.

\paragraph{On-policy alignment on target language} Finally, in the third stage, the language model is aligned to respond in a helpful, truthful and safe way -- without losing the fluency it acquired during the initial pretraining stage. This is achieved by training with online on-policy reinforcement learning where the model is trained solely on responses sampled from itself. In this way, the model is never pushed away from the subspace of fluent outputs it has learned to prefer during pretraining. A key observation is that we do not need to train any reward model as it suffices to use an LLM-as-a-judge system to provide the reward signal \citep{10.5555/3666122.3668142}; \textit{as long as the judge understands the target language, it does not have to be fluent to produce a fluent policy.} We evaluate this hypothesis later in \Cref{sec:experiment-2}. This effectively allows us to bootstrap fluent models in languages without any instruction datasets and without any existing fluent language models.

\subsection{Online on-policy reinforcement learning}
\label{sec:rl}

We now describe the components of our on-policy training: the reward signal, the policy gradient objective, KL regularization, and the distributed implementation. It is important to note that any implementation of online on-policy training with feedback from an LLM-as-a-judge system should work comparably well in terms of fluency of the final policy network -- our implementation is chosen to be simple and comparable to the baseline approach of supervised finetuning, as detailed later in \Cref{sec:experiment-1}. The algorithm does not directly optimize fluency: as detailed below, the reward signal captures the overall quality of responses based on the standard criteria such as correctness, safety or conciseness -- without explicitly targeting fluency (\Cref{app:prompts}).

\paragraph{Reward model} The standard approach to on-policy alignment -- reinforcement learning from human feedback \citep[RLHF;][]{christiano2017deep, stiennon2020learning, 10.5555/3600270.3602281} -- first trains a Bradley-Terry reward model on a preference dataset from the target domain. Since we are restricted to a lower-resource language, we assume that such a dataset is not available -- instead, we rely on direct reinforcement learning from AI feedback \citep[d-RLAIF;][]{10.5555/3692070.3693141}.

In this scenario, we only use the domain knowledge to create a prompt template (constitution) that clearly guides a multilingual language model to judge the quality of responses. As evident from later results in \Cref{sec:experiment-2}, this setup provides enough signal that even a disfluent judge can train a fluent and capable policy -- as long as the judge has some level of understanding of the target language.

\paragraph{Objective function} Let us start with defining the objective $\mathcal{J}$ to maximize during this post-training stage. The objective in \Cref{eq:rl-objective} states that we want to find parameters $\bm\theta$ of our language model $\pi_{\bm\theta}$ (the policy model) that maximize the reward $r$ given to a prompt $\bm{x}=(x_1, x_2, \dots x_{|\bm{x}|})$ and its corresponding response $\bm{y}=(y_1, y_2, \dots y_{|\bm{y}|})$ sampled from the policy $\pi_{\bm\theta}$:
\begin{equation}
    \mathop{\text{argmax}}_{\bm\theta} \mathcal{J}(\bm{\theta}) \defeq \mathop{\text{argmax}}_{\bm\theta}\mathop{\mathbb{E}}_{\bm{x}\, \sim\,\mathcal{D},\, \bm{y}\, \sim\,\pi_{\bm\theta}(\cdot|\bm{x})}r(\bm{x}, \bm{y}).
\label{eq:rl-objective}
\end{equation}

Following the majority of works on LM alignment, we optimize the objective with \emph{policy gradient} methods \citep{10.1007/BF00992696, sutton2000policy} that perform gradient descent on $-\nabla_{\!{\bm\theta}}\mathcal{J}(\bm{\theta})$ using online on-policy samples $\bm{y}$ from the policy model $\pi_{\bm\theta}$:
\begin{equation}
    -\nabla_{\!{\bm\theta}}\mathcal{J}(\bm{\theta}) = -\mathop{\mathbb{E}}_{\bm{x}\, \sim\,\mathcal{D},\,\bm{y}\, \sim\,\pi_{\bm\theta}(\cdot|\bm{x})} r(\bm{x}, \bm{y})\, \nabla_{\!\bm\theta}\log \pi_{\bm\theta}(\bm{y}|\bm{x}).
\label{eq:reinforce}
\end{equation}

Directly using \Cref{eq:reinforce} for training gives us the \textsc{Reinforce} algorithm \citep{10.1007/BF00992696}. To increase its convergence speed and stability, we modify the rewards and optimize advantages $A(\bm{x}, \bm{y})$ instead of $r(\bm{x}, \bm{y})$. Following \textsc{Reinforce with baseline} \citep{10.5555/2074022.2074088}, we subtract the baseline score $b(\bm{x})$, and following \cite{karpathy2016pong}, we further normalize by the dispersion factor $s(\bm{x})$; giving us $\smash{A(\bm{x},\bm{y})\defeq \nicefrac{(r(\bm{x}, \bm{y}) - b(\bm{x}))}{s(\bm{x})}}$. While these additional factors are often estimated by separately trained models \citep{konda2000actor, schulman2017proximalpolicyoptimizationalgorithms, christiano2017deep}, we choose a more straightforward approach and estimate them as the sample mean and the sample standard deviation over $G$ responses \citep{Kool2019Buy4R, ahmadian-etal-2024-back, shao2024deepseekmathpushinglimitsmathematical}:
\begin{equation}
    \hat{A}(\bm{x}, \bm{y}) \defeq \frac{r(\bm{x}, \bm{y}) - \text{mean}\bigl(\{r(\bm{x}, \bm{y}^{(i)}){\}}_{i=1}^{\vphantom{1}G}\bigr)}{\text{std}\bigl(\{r(\bm{x}, \bm{y}^{(i)}{\}}_{i=1}^{\vphantom{1}G}\bigr)}.
\label{eq:advantage}
\end{equation}

\strut Putting \Cref{eq:reinforce} and \Cref{eq:advantage} together, we can define the loss function $\mathcal{L}(\bm\theta, \bm{x})$ of a single query $\bm{x}$ for optimizing the parameters $\bm\theta$ of a policy $\pi_{\bm\theta}$. One important detail is to account for the potential length bias \citep{liu2025understandingr1zeroliketrainingcritical} when computing the log-likelihood $\log \pi_{\bm\theta}(\bm{y}|\bm{x})$ as $\smash{\sum_{j=1}^{|\bm{y}|} \log \pi_{\bm\theta}(y_{j}|\bm{x},\bm{y}_{{<j}})}$ and normalizing by $\nicefrac{1}{|\bm{y}|}$. As proposed by \cite{journals/corr/abs-2506-10910}, the length bias can be mitigated by dividing by the total length of responses $\smash{\{\bm{y}^{(i)}\}_{i=1}^{\vphantom{1}G}}$ to a prompt $\bm{x}$. Then the token-level loss function becomes:
\begin{equation}
    \nabla_{\!\bm\theta}\mathcal{L}(\bm\theta) \defeq - \mathop{\mathbb{E}}_{\bm{x}\, \sim\,\mathcal{D}, \{\bm{y}^{(i)}\!{\}\!}_{i=1}^{\vphantom{1}G}\, \sim\,\pi_{\bm\theta}(\cdot|\bm{x})} \Biggl[ \frac{1}{\sum_i{\!|\bm{y}^{(i)}|}}\sum_{i=1}^G \hat{A}(\bm{x}, \bm{y}^{(i)}) \sum_{j=1}^{|\bm{y}^{(i)}|} \nabla_{\!\bm\theta}\log \pi_{\bm\theta}(y^{(i)}_j|\bm{x},\bm{y}^{(i)}_{<j})\Biggr],
\label{eq:policy_loss}
\end{equation}

Note that unlike most current work on LM alignment, we do not modify the loss function to account for sampling from a proximal policy (by clipping outlier samples and importance sampling) as in PPO \citep{schulman2017proximalpolicyoptimizationalgorithms}. As detailed below, the synchronous parallelization makes all samples almost fully on-policy and we have found no benefit in moving away from the simple \textsc{Reinforce}-like loss function in \Cref{eq:policy_loss}. 

\paragraph{KL-divergence regularization} A well known issue of policy gradient methods in RLHF is that they optimize model-based rewards instead of the true (and unknown) rewards. When the policy is modeled by a large language model with billions of parameters, it can very quickly find shortcuts in the reward model and \textit{reward-hack} its measured performance while degrading its true performance. The most common way to mitigate this issue is to add a soft constraint to the optimization problem that pushes the policy to stay close to its original state -- the distance is usually measured by the KL divergence and the strength of the constraint is parametrized by $\beta$ \citep{kullback1951information, jaques2019way}:
\begin{equation}
    \mathop{\text{argmax}}_{\bm\theta} \mathcal{J}(\bm{\theta}) \defeq \mathop{\text{argmax}}_{\bm\theta}\mathop{\mathbb{E}}_{\bm{x}\, \sim\,\mathcal{D},\, \bm{y}\, \sim\,\pi_{\bm\theta}(\cdot\mid \bm{x})} \Bigl[ r(\bm{x}, \bm{y}) - \beta D_\text{KL}[\pi_{\bm\theta}(\cdot\mid \bm{x})\,||\,\pi_{\bm\theta_{\text{ref}}}(\cdot\mid \bm{x})] \Bigr].
\label{eq:kl-objective}
\end{equation}

In order to optimize the objective $\mathcal{J}$ according to the new definition instead of the simpler one from \Cref{eq:rl-objective}, we need to introduce an additional loss term $\mathcal{L}_{KL}$ (weighted by $\beta$) that will push the trained policy $\pi_{\bm\theta}$ closer to the output distribution $\pi_{\bm\theta_{\text{ref}}}$.

The problem with KL divergence is that its exact computation is intractable in most cases. This means that we have to approximate it in practice, the most common way is to simply use the already sampled prompts with responses and compute a direct Monte-Carlo estimate of $\smash{\mathop{\mathbb{E}}_{\bm{x}\, \sim\,\mathcal{D},\, \bm{y}\, \sim\,\pi_{\bm\theta}(\cdot\mid \bm{x})}\bigl[\text{log}\,\nicefrac{\pi_{\bm\theta}(\bm{y}|\bm{x})}{\pi_{\bm\theta_{\text{ref}}}(\bm{y}|\bm{x})}\bigr]}$, as done in the seminal RLHF work by \cite{stiennon2020learning}. While straightforward, this approximation is very rough and ill-behaved, even becoming negative sometimes.

These estimates use only a small fraction of information available in the probability distributions given by $\pi_{\bm{\theta}}(\cdot|\bm{x}, \bm{y}_{<i})$ -- only the single value of $\pi_{\bm{\theta}}(y_i|\bm{x}, \bm{y}_{<i})$. Instead, we can get provably tighter estimates when we Rao-Blackwellize the Monte-Carlo estimation by using the full next-token distributions over the vocabulary $\mathcal{V}$. \cite{amini2025betterestimationkldivergence} prove that this estimation is unbiased and has lower variance than the standard Monte-Carlo estimation.
\begin{equation}
\mathcal{L}_{KL}(\bm\theta) \defeq \mathop{\mathbb{E}}_{\bm{x}\, \sim\,\mathcal{D}, \bm{y}\, \sim\,\pi_{\bm\theta}(\cdot\mid \bm{x})}\Biggl[ \sum_{i=1}^{|\bm{y}|} \underset{\substack{\\[-1ex] \underbrace{\hspace{20em}} \\[1ex] D_\text{KL}[\pi_{\bm\theta}(\cdot\,|\,\bm{x}, \bm{y}_{<i})\,||\,\pi_{\bm\theta_{\text{ref}}}(\cdot\,|\,\bm{x}, \bm{y}_{<i})]}}{\sum_{w=1}^{|\mathcal{V}|}\pi_{\bm\theta}(y_i=w\,|\,\bm{x}, \bm{y}_{<i})\cdot\text{log}\frac{\pi_{\bm\theta}(y_i=w\,|\,\bm{x}, \bm{y}_{<i})}{\pi_{\bm\theta_\text{ref}}(y_i=w\,|\,\bm{x}, \bm{y}_{<i})}}\Biggr].
\label{eq:approx-kl}
\end{equation}

The computational overhead of the Rao-Blackwellized estimate is negligible because it still requires only a single forward-backward pass through the policy model that is done even without any KL regularization. Another benefit of regularizing the full output distribution is that it eliminates the need for another loss term for regulating the output entropy -- as used in most RLHF works -- further simplifying the training method.

\begin{figure}[h!]
    \centering
    \includegraphics[width=0.8\linewidth]{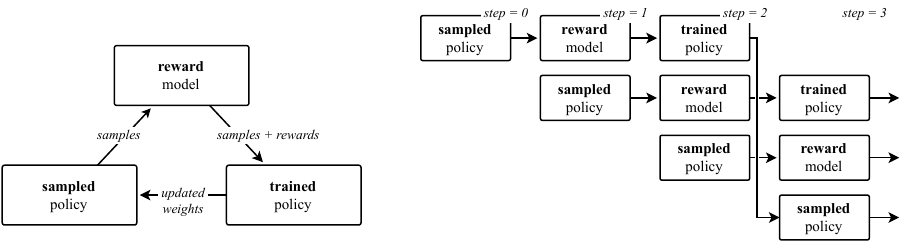}
    \caption{\textbf{Left: Reinforcement-learning cycle.} This diagram demonstrates the sequential nature of online RL training: each training step starts by sampling new responses from the policy model, followed by sampling response-judgments from the reward model, and then updating the weights of the policy model based on the sampled responses and rewards. \textbf{Right: Parallelization.} Breaking the cycle and postponing the update of the sampled policy allows for running all three models at the same time (vertically-aligned blocks are ran concurrently on different GPU nodes).}
    \label{fig:rl-cycle}
\end{figure}

\paragraph{Distributed setup} As opposed to supervised finetuning, RL approaches need several language models to be fully materialized and used at the same time (the trained policy, the reference policy, the sampled policy and the reward model). In principle, these models should be run sequentially in a cycle (\Cref{fig:rl-cycle}), but that is inefficient in practice and the cycle needs to be broken and parallelized. As illustrated below in \Cref{fig:rl-cycle}, this can be achieved by postponing the update of the sampled policy -- effectively turning the training slightly off-policy.

Unlike other approaches to distributed RL \citep{48558, noukhovitch2025faster, journals/corr/abs-2506-10910}, our parallelization is completely synchronous. This can make the resources allocated for sampling underutilized (all workers have to wait until the longest response completes), but that does not impact the overall efficiency much as most resources are allocated to the reward models. On the other hand, our samples are unbiased (asynchronous approaches typically up-sample problems with short responses), and the completely synchronous training cycle simplifies the implementation, as well as the objective function. Since the samples are guaranteed to be off-policy by just three steps, we can still rely on on-policy training techniques without having to resort to more complicated and less stable proximal-policy methods such as PPO \citep{schulman2017proximalpolicyoptimizationalgorithms}.

\section{Experiment: Online on-policy training maintains fluency}
\label{sec:experiment-1}

The main experiment of this paper aims to answer the central question: \textit{Does online on-policy training produce more fluent language models than supervised finetuning on translated data?} To answer this, we designed the experiment to make on-policy training as similar to supervised finetuning as possible -- using the same base model, the same training data, and the same number of training samples. As a case study for models trained on lower-resource language, we trained all models on Norwegian Bokmål; then we asked five native speakers to do pair-wise fluency comparisons of outputs generated from these models.

In total, we compare three post-training approaches represented by three language models that are based on the same pretrained model, as illustrated in \Cref{fig:genealogy}:
\begin{figure}[h]
    \centering
    \includegraphics[width=1.0\linewidth]{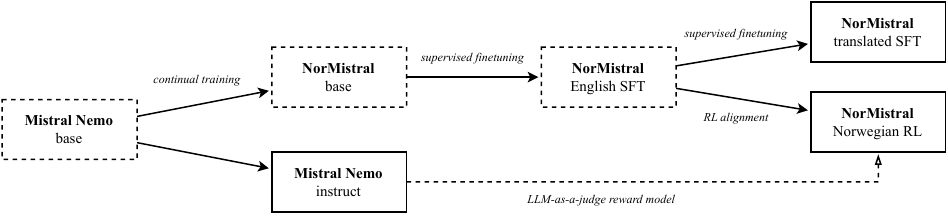}
    \caption{\textbf{Genealogy of the compared models.} The three models compared in the main fluency experiment (highlighted in bold boxes) all originate from a single base model -- Mistral Nemo 12B (left). The official instruction-tuned version of this model also serves as the judge for our reinforcement-learning approach.}
    \label{fig:genealogy}
\end{figure}

\paragraph{\textit{Approach 1}: Norwegian RL} We follow the method described in \Cref{sec:rl} when training this model. More specifically, we start from \texttt{NorMistral 11B}, a continually pretrained Norwegian base model from \cite{samuel-etal-2025-small}, and supervise-finetune it for a single epoch on the 1\,000 English conversations from LIMA \citep{10.5555/3666122.3668522}. Then, in the final reinforcement-learning stage, we use the same dataset as the second approach: No Robots by \cite{no_robots}. In each training step, we take 128 translated prompts from this dataset, sample a group of 8 responses for each prompt. Each response is then graded by a judge that also has access to the gold English response provided in the No Robots dataset -- the judge prompt itself is attached in \Cref{app:prompts}. In order to rule out the possibility that the policy acquires its fluency from the judge model, we use \texttt{Mistral Nemo 12B} as the judge -- a model with the same origin as the policy itself that is also evaluated in this experiment as the approach number three. More technical training details are described in \Cref{app:hyperparameters}. Many of these choices are ablated later in \Cref{sec:further-experiments} -- for example, it is clear that Mistral Nemo is a poor reward model compared to larger language models, which should not, however, influence the fluency of the trained policy (\Cref{sec:experiment-2}).

\paragraph{\textit{Approach 2}: Translated SFT} As illustrated in \Cref{fig:genealogy}, the translated-SFT model is initialized from the same checkpoint as the first approach. Then we also finetune it on the same machine-translated No Robots dataset -- however in this scenario, we directly finetune the model on the translated responses by minimizing their conditional negative log-likelihood (in the first approach, the non-translated responses are only used as hints for the reward model). To translate the full No Robots dataset to Norwegian Bokmål, we use the \texttt{Unbabel/Tower-Plus-72B} model -- currently the state-of-the-art open-weights translation model with explicit support for Norwegian \citep{rei2025towerplus}. Other strong translation models are ablated later in \Cref{sec:experiment-5}.

\paragraph{\textit{Approach 3}: Mistral Nemo} The last tested model is \texttt{Mistral-Nemo-Instruct-2407} by \cite{mistralai2024nemo}. Including it in this experiment serves three purposes: firstly, to give a reference baseline from an externally trained model; secondly, to test the fluency of the reward model that has been used in the first approach; and thirdly, to evaluate the performance of the standard multilingual -- but English-focused -- post-training that has been used in most major releases of the latest language models. This model shares the same origin as the previous two. However, while its weights are openly available, no details about its post-training process have been published.

\paragraph{Manual annotation of generated outputs} Fluency is a language feature that is difficult to accurately measure using statistical models, but it should be relatively straightforward to judge for native speakers of that language. We therefore hired five research assistants, all native Norwegian speakers, to compare the fluency of responses generated for a pool of prompts. The prompts are gathered from the Norwegian Bokmål \texttt{mimir-instruct} dataset from \cite{rosa-etal-2025-impact}, which consists of artificially generated responses to prompts written by native Norwegian speakers. Since we use this data primarily as seeds for diverse generated outputs to evaluate for fluency, we select the 100 queries with the longest gold responses as the seed prompts; to ensure sufficiently long responses for meaningful fluency comparison. Then we sample a single response from each of the three evaluated models -- using Nucleus sampling with temperature of $0.5$, \texttt{top\_k} of 64 and \texttt{top\_p} of 0.9 \citep{Holtzman2020The}. The responses are formed into 3 pairs per prompt (one for each model pair), yielding 300 pairs in total. These pairs are then rated by the annotators in the A/B testing fashion -- \emph{Which is more fluent: response A, response B, or are they equally fluent?} Each annotator is presented with the 300 pairs of responses in a randomized order; going through the full set of pairs took each annotator roughly 15--20 hours. We give more details about the actual annotation guidelines and the overall process in \Cref{app:annotation}. The resulting dataset (with anonymized annotator names) is published online at {\small\url{https://hf.co/datasets/ltg/normistral-fluency-annotation}}.

\paragraph{Results} We show the resulting win-rates from the manual annotation in \Cref{tab:winrates}. These win-rates are calculated by going through all pair-wise comparisons and either giving the winning model (preferred by more annotators) a single point or giving both models half a point in case of a tie -- so-called $1/0.5/0$ method of aggregation \citep{Copeland1951}.

\renewcommand{\arraystretch}{1.75}

\begin{table}[h!]
\centering
\small
\caption{\textbf{Model win-rates according to the manual fluency annotation.} The values show the win-rate percentages of the row-wise models over the column-wise models. The last column aggregates the win-rates of each model.\vspace{-0.5em}}
\begin{tabular}{@{}l@{\hspace{2em}}ccc@{\hspace{4em}}c}
\toprule
\textbf{Model} & \textbf{on-policy RL} & \textbf{translated SFT} & \textbf{Mistral Nemo} & \hspace{1.25em}\textbf{average}\hspace{1.25em} \\
\midrule
on-policy RL & --- & \cellcolor[rgb]{0.703,0.802,0.984} 67.5 & \cellcolor[rgb]{0.437,0.572,0.953} 91.8 & \cellcolor[rgb]{0.570,0.704,0.998} 79.7 \\
translated SFT & \cellcolor[rgb]{0.964,0.754,0.656} 32.5 & --- & \cellcolor[rgb]{0.483,0.622,0.975} 87.5 & \cellcolor[rgb]{0.777,0.841,0.947} 60.0 \\
Mistral Nemo & \cellcolor[rgb]{0.908,0.458,0.357} 8.2 & \cellcolor[rgb]{0.933,0.520,0.407} 12.5 & --- & \cellcolor[rgb]{0.921,0.489,0.381} 10.3 \\
\bottomrule
\end{tabular}
\label{tab:winrates}
\end{table}

The results clearly show that the most-preferred responses are from the on-policy training method, followed by translated SFT and then by Mistral Nemo. This supports our main claim since it demonstrates that the policy can indeed outperform its judge on fluency. It is worth noting that the first two approaches produced very fluent outputs and the fluency of $\nicefrac{1}{3}$ of their pair-wise comparisons was agreed to be `equal' -- based on qualitative inspection, the difference between them mostly stems from infrequent traces of translationese in the SFT outputs (\Cref{app:samples}).

\section{Further evaluations and ablations}
\label{sec:further-experiments}

This section studies the effect of various post-training choices on the final performance in more detail. We scale up the manual fluency scoring from \Cref{sec:experiment-1} by introducing an automatic fluency estimate, and also focus on more general performance of the trained models by incorporating Norwegian understanding and generation benchmarks.

\subsection{Evaluation metrics}

\paragraph{Automatic fluency evaluation} We have to rely on model-based measurement of fluency to lower the cost of evaluating all experiments in this section. The Norwegian fluency model is trained like a standard Bradley-Terry reward model \citep{stiennon2020learning} on a dataset $\mathcal{D}$ of paired preferred and non-preferred texts $(\bm{x}_w, \bm{x}_l) \in \mathcal{D}$. Specifically, we add a scalar linear head to a pretrained language model (\texttt{NorMistral 11B} in our case) and finetune it by minimizing the following loss:
\begin{equation}
    \mathcal{L}_{fluency}(\bm{\theta}) \defeq -\mathop{\mathbb{E}}_{\bm{x}_w, \bm{x}_l\,\sim\,\mathcal{D}}\,\Bigl[\log\sigma\bigl(r_{\bm\theta}(\bm{x}_w) - r_{\bm\theta}(\bm{x}_l)\bigr)\Bigr].
\end{equation}

We create the training data by combining existing resources and newly synthesized texts. Firstly, we use the Norwegian ASK-GEC corpus of corrected language-learner essays \citep{jentoft2023grammatical} -- from this corpus, we take all sentences with mistakes as the non-preferred texts and their (partially) corrected versions as the preferred texts. Secondly, we perform backtranslation with the OPUS collection of Norwegian-English machine translation models \citep{Tiedemann2023DemocratizingNM}: we sample a clean Norwegian sentence from the Norwegian Dependency Treebank \citep[NDT;][]{solberg-etal-2014-norwegian} or the Norwegian Review Corpus \citep[NoReC;][]{velldal-etal-2018-norec}, then we sample one model to translate the sentence to English and then sample another model to translate it back to Norwegian; finally, the original sentence is cast as preferred and the backtranslated version as non-preferred. The ASK-GEC corpus trains the fluency scorer to take grammaticality into account while the second synthetic source focuses on translationese and lexical issues.

The fluency scorer can be directly evaluated by utilizing the manual annotations from the previous \Cref{sec:experiment-1}. Looking at all instances where the annotators agreed that one response is preferred over another response, the fluency scores agree with this ranking in 85.5\% of cases. This agreement is even slightly higher than the agreement among annotators -- when limiting their annotation to non-ties (for comparability), they agree with the consensus in 83.2\% of cases, which highlights how subjective the notion of fluency can be. To lower the variance of the fluency score, we sample 16 responses from each evaluated model and average the scores. The raw scores are sigmoid-normalized into percentage values for clarity. When applied to the three approaches from the previous section, the fluency scores are 2.47 (92.2\%) for \textit{on-policy RL}, 1.94 (85.7\%) for \textit{translated SFT}, and 0.76 (65.3\%) for \textit{Mistral Nemo}, which correspond to the manual evaluation of these three models. 

\paragraph{Natural language understanding (NLU) evaluation} We use the native-Norwegian reading-comprehension task NorQuAD to assess the level of Norwegian language understanding \citep{ivanova-etal-2023-norquad}. In order to account for the variable and conversational outputs of instruction-tuned models, we evaluate the correctness of each generated response with an extra call to a judge, \texttt{Llama 3.3 70B} \citep{Grattafiori2024llama}, that compares it against the gold answer. This ensures that the evaluation is invariant to formatting variation of the generated outputs. We then report the average accuracy as the approximate Norwegian NLU score.

\paragraph{Natural language generation (NLG) evaluation} In order to assess the Norwegian generative abilities, we use two Norwegian benchmarks that were designed specifically for this purpose:  NorRewrite and NorSummarize \citep{mikhailov-etal-2025-noreval}; these benchmarks test the instruction-following, creative writing, and summarization abilities of Norwegian language models. 
We follow the original implementation of these benchmarks, which evaluates the quality of each response by comparing it pairwise with another response and automatically judging it with \texttt{Llama 3.3 70B}. We report the win-rate percentages of individual models against the smallest evaluated model, \texttt{Llama 3.1 8B} \citep{Grattafiori2024llama}.

\subsection{Fluent policy does not need a fluent reward model}
\label{sec:experiment-2}

The main experiment has shown that even a disfluent judge (Mistral Nemo 12B) can produce a policy that is substantially more fluent than the judge itself. In this section, we investigate this phenomenon more thoroughly by checking that it is not an anomaly and that it holds for a diverse range of judges. First, we evaluate each judge on the three benchmarks described above: natural language understanding (NLU), generation (NLG), and fluency. Then, using a judge to provide the reward signal, we train a policy with the same method as in \Cref{sec:experiment-1}, and evaluate the policy on fluency -- to assess the effect of the choice of judge.

\renewcommand{\arraystretch}{1.35}

\begin{table}[h!]
\centering
\small
\caption{\textbf{The effect of the judge's knowledge of Norwegian on the trained policy.} The table shows the average Norwegian understanding, generation and fluency scores (\Cref{sec:further-experiments}) for different judges, and the fluency of the policy models trained with reward signals from these judges. The fluency scores are color-coded so that disfluent models are red and fluent models are blue.\vspace{-0.5em}}

\begin{tabular}{@{}lcccccc}
\toprule
\textbf{} &  & \multicolumn{3}{c}{\textbf{Judge performance}} &  & \hspace{-1em}\textbf{Fluency of}\hspace{-1em} \\
\textbf{Judge} &  & \textbf{NLU} & \textbf{NLG} & \textbf{Fluency} & \hspace{1em} & \hspace{-1em}\textbf{trained policy}\hspace{-1em} \\ \midrule
Mistral Nemo 12B &  & 87.5 & 29.7 & \cellcolor[rgb]{0.911,0.840,0.800} 67.0 &  & \cellcolor[rgb]{0.522,0.660,0.989} 92.2 \\
Mistral Large 123B &  & 90.0 & 70.4 & \cellcolor[rgb]{0.674,0.784,0.992} 83.4 &  & \cellcolor[rgb]{0.483,0.622,0.975} \textbf{94.2} \\
Mixtral 8x22B &  & 91.3 & 20.2 & \cellcolor[rgb]{0.862,0.865,0.869} 70.9 &  & \cellcolor[rgb]{0.524,0.662,0.989} 92.1 \\
Llama 3.1 8B &  & 86.4 & 50.0 & \cellcolor[rgb]{0.948,0.796,0.718} 62.8 &  & \cellcolor[rgb]{0.509,0.647,0.985} 92.9 \\
Llama 3.3 70B &  & 90.7 & 57.7 & \cellcolor[rgb]{0.661,0.775,0.995} \textbf{84.2} &  & \cellcolor[rgb]{0.499,0.638,0.981} 93.5 \\
Qwen 2.5 14B &  & 89.6 & 43.5 & \cellcolor[rgb]{0.843,0.328,0.268} 39.0 &  & \cellcolor[rgb]{0.506,0.645,0.984} 93.1 \\
Qwen 2.5 32B &  & 91.7 & 59.9 & \cellcolor[rgb]{0.897,0.434,0.339} 43.2 &  & \cellcolor[rgb]{0.493,0.631,0.979} 93.9 \\
Qwen 2.5 72B &  & \textbf{92.0} & \textbf{75.2} & \cellcolor[rgb]{0.958,0.600,0.479} 50.7 &  & \cellcolor[rgb]{0.509,0.647,0.985} 92.9 \\
\bottomrule
\end{tabular}
\label{tab:judge}
\end{table}

\paragraph{Results} We have evaluated language models of different sizes and different levels of Norwegian knowledge -- three Mistral models \citep{mistralai2024nemo, jiang2024mixtralexperts}, three Qwen models \citep{qwen2025qwen25technicalreport}, and two Llama models \citep{Grattafiori2024llama} -- the results of the evaluation are shown in \Cref{tab:judge}. Comparing the fluency scores of judges with policies, there is no apparent correspondence (the Pearson correlation coefficient is $0.067$); \textit{the policies are fluent regardless of the (dis)fluency of their judge.} There is also no clear relation between the other two measures of judge quality on the resulting fluency. We hypothesize that fluency remains stable because the policy is trained exclusively on its own samples, which are fluent thanks to targeted pretraining. The choice of judge model appears to affect other response qualities but not fluency.

\subsection{The effect of training length on fluency}
\label{sec:experiment-4}

We further validate the claim that fluency is consistently stable for on-policy training by looking at the change in fluency score throughout training. 

\begin{figure}[h!]
    \centering
    \includegraphics[width=\linewidth]{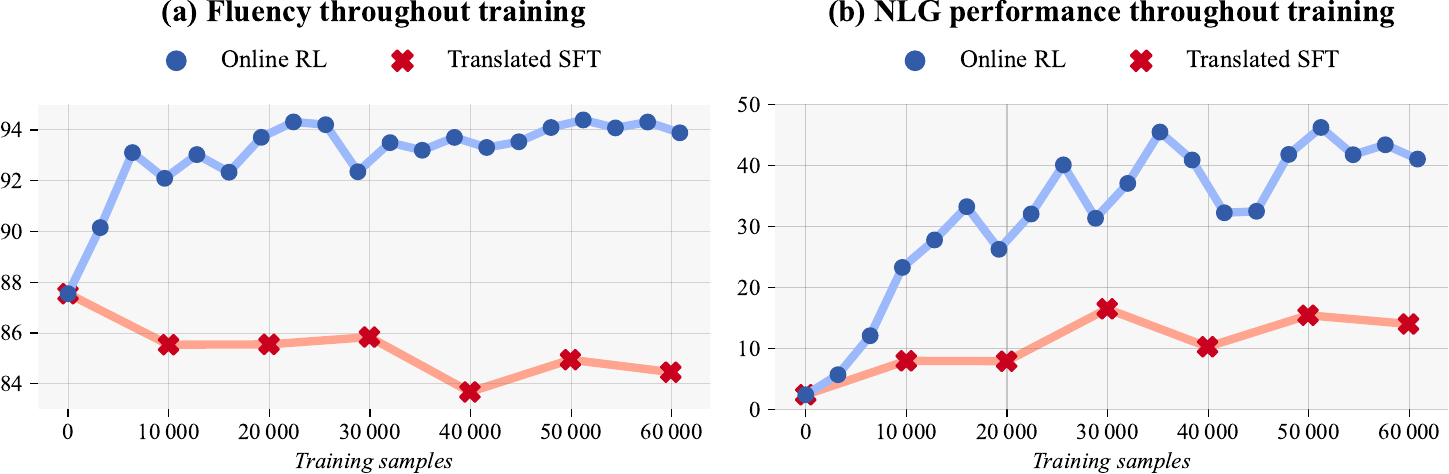}
    \caption{\textbf{Fluency and NLG scores throughout training.} We measure the performance score every 25 training steps for the reinforcement learning (in blue) and every epoch for the SFT training.}
    \label{fig:fluency-convergence}
\end{figure}

\paragraph{Results}\Cref{fig:fluency-convergence} shows the result of training a policy supervised by Mistral Large and evaluating its fluency and NLG performance every 25 training steps. \textit{The fluency score remains stable around 93\% after the initial convergence in the first 50 training steps.} Upon closer inspection, the lower initial fluency score of 87.5\% can be explained by the increased likelihood of responding in English (not by disfluent Norwegian per se) triggered by the previous English SFT stage; then the model learns to consistently respond in Norwegian, which leads to the apparent increase in fluency {(\color{blue}see \Cref{app:samples})}. On the other hand, when looking at the change of fluency throughout SFT training on translated Norwegian, we can see a clear decrease in fluency from the initial starting point, which can only be attributed to responding with consistent -- but slightly disfluent -- Norwegian. The longer reinforcement-learning run also allows us to look at the development of the directly optimized training metrics; these are appended in \Cref{app:training-metrics}.

\subsection{The importance of not training the policy on any translated responses}
\label{sec:experiment-3}

The previous experiment has highlighted the importance of the initial SFT stage and the ability of the policy model to recover from the sudden shift to English. As described earlier, we only train on the 1\,000 examples (31 training steps) from the LIMA dataset \citep{10.5555/3666122.3668522} in the initial SFT stage. The results in \Cref{tab:sft-length} suggest that training on more samples can slightly degrade fluency. Even clearer fluency deterioration is seen after machine-translating the dataset from English -- \emph{even a small amount of translated data can introduce a measurable amount of disfluencies in the final policy.}

\subsection{Impact of translation quality on SFT performance}
\label{sec:experiment-5}

Even though we have selected Tower-Plus as the state-of-the-art model for translation from English to Norwegian \citep{rei2025towerplus}, a possible explanation of our results is that the observed disfluencies are specific to this particular translation model rather than a general phenomenon. Thus, we translate the No Robots dataset with several alternative models, finetune on each translation, and compare the resulting fluency. We only test translation models that explicitly support English and Norwegian, which includes LLM-based translation models, massively multilingual models, and a traditional small model trained specifically for translation. The fluency scores in \Cref{tab:translation} confirm that Tower-Plus, as the largest available translation model, results in the most fluent policy.

\renewcommand{\arraystretch}{1.2}
\begin{table}[h!]
\centering
\begin{minipage}{0.45\textwidth}
      \centering
      \small
      \caption{\textbf{The effect of the initial SFT stage.} We ablate various settings and measure the final fluency when starting reinforcement learning from such SFT checkpoints.\vspace{0.0em}}
      \begin{tabular}{@{}lc@{}}
            \toprule
            \textbf{SFT settings}                    & \textbf{RL fluency} \\ \midrule
            English dataset (1 epoch)    &    \textbf{94.2}   \\
            English dataset (2 epochs)   &    93.2  \\
            English dataset (4 epochs)   &    92.8     \\
            Translated dataset (1 epoch) &   91.0         \\ \bottomrule
        \end{tabular}
        \label{tab:sft-length}
\end{minipage}
\hfill
\begin{minipage}{0.5\textwidth}
      \centering
      \small
        \caption{\textbf{The effect of using different machine-translation models.} We train models on the translated dataset and measure their fluency.}
        \begin{tabular}{@{}lcc@{}}
            \toprule
            \textbf{Translation model}                    & \textbf{Size} & \textbf{Fluency}    \\ \midrule
            Tower-Plus {\tiny\citep{rei2025towerplus}}   & 72.7B &   \textbf{85.7}                   \\
            MADLAD-400 {\tiny\citep{10.5555/3666122.3669062}} & 10.7B & 82.4                 \\

            Seed-X {\tiny\citep{cheng2025seedxbuildingstrongmultilingual}}   & 7.5B &  73.4                 \\
            NLLB-200 {\tiny\citep{nllbteam2022languageleftbehindscaling}}   &  3.3B & 75.5                  \\
            OPUS Eng-Gem {\tiny\citep{Tiedemann2023DemocratizingNM}}              & 0.1B & 68.2                 \\ \bottomrule
        \end{tabular}
        \label{tab:translation}
\end{minipage}
\end{table}

\section{Related work}

\paragraph{Post-training on lower-resource languages} As mentioned above, there have been numerous works that focused on supervised finetuning of language models for lower-resource languages \citep[\textit{inter alia}]{10386605, pipatanakul2023typhoon}, substantially less attention has been paid to reinforcement learning for preference optimization in such languages \citep{lai-etal-2023-okapi, dang-etal-2024-rlhf}. While fluency in less-spoken languages is a significant limitation of current language models, there has not been much interest from the research community in this topic -- likely because accurate fluency evaluation is difficult. \cite{dang-etal-2024-rlhf} consider fluency as an important aspect of multilingual performance, but only include it as part of a general LLM-as-a-judge evaluation prompt, whose accuracy is not validated. \cite{zhang2025lingualifteffectivetwostageinstruction} use human annotators to evaluate fluency of their models, but only after first back-translating the generated responses to English. A recent paper by \cite{sainz2025instructinglargelanguagemodels} focuses on the effect of post-training configuration on fluency, similarly to our work -- they have fluency assessed by native (Basque) speakers, but only consider the effect of different data mixtures for supervised finetuning.

\paragraph{Translationese and translation artifacts} The phenomenon of translationese \citep{475831, cb567c06c5bd4728a25f5b5cead318be, alma991004198289705502} is well-documented in both human and machine translation \citep{koppel-ordan-2011-translationese, 10.1093/llc/fqt031, vanmassenhove-etal-2019-lost, bizzoni-etal-2020-human}. \cite{wang-etal-2023-understanding} have demonstrated that translationese in training data impacts performance in cross-lingual summarization, with models trained on translated data producing outputs that differ systematically from those trained on original text. This finding motivates our approach: rather than trying to mitigate translationese artifacts during training, we avoid them entirely by never training on any translated text. The concurrent work by \cite{kunz2026preferencesidiomaticlanguageacquired} studies the effect of fine-tuning on machine-translated Swedish and, similarly to us, found that it leads to a rapid loss of fluency.

\paragraph{Weak-to-strong generalization}
The weak-to-strong generalization framework demonstrates that strong models consistently outperform their weak supervisors when fine-tuned on the labels generated by the weak supervisors \citep{10.5555/3692070.3692266}. Following these empirical findings, \cite{NEURIPS2024_e4a0d8ae} has proposed a theoretical frameworks that explains this phenomenon. It is particularly relevant to our setting: just as strong models can exceed the capabilities of weak supervisors, our Norwegian-pretrained models can achieve fluency beyond what the quality of the supervision signal would suggest.

\paragraph{RLAIF and the judge-policy disconnect} The distinction between the capabilities of the reward model and the policy model has been explored in the RLAIF literature. \cite{10.5555/3692070.3693141} discovered that RLAIF can achieve comparable performance to RLHF even when the LLM judge is not strong enough, which is similar to our finding that disfluent judges can guide fluent policy models. More generally, the disconnect between evaluation and performance is well-documented: LLM judges can accurately predict human preferences even when they cannot generate equally high-quality outputs themselves \citep{10.5555/3666122.3668142}. The key insight is that evaluation is often a simpler, more focused task than generation -- a principle that explains why disfluent judges can still provide useful training signals for Norwegian responses. Furthermore, recent work on self-rewarding models shows that LLMs can iteratively improve through self-evaluation, further demonstrating that the ability to discriminate quality does not require the ability to generate at that quality level \citep{10.5555/3692070.3694459}. This asymmetry between recognition and generation capabilities suggests why our approach works: the judge models only need to identify responses that better satisfy the task requirements; they do not need to generate fluent Norwegian themselves.

\paragraph{Cross-lingual transfer and unsupervised translation} Our work can be connected to several lines of research on learning from imperfect supervision. The literature on zero-shot cross-lingual transfer has extensively studied how models trained on one language can perform in other languages without direct supervision. \cite{pires-etal-2019-multilingual} have demonstrated that multilingual BERT models fine-tuned on one language can achieve reasonable zero-shot performance on tasks in other language. More recent work has shown that this transfer can be improved through better pretraining \citep{conneau-etal-2020-unsupervised, xue-etal-2021-mt5} and cross-lingual alignment techniques \citep{artetxe-etal-2020-cross}.

\section{Conclusion}

In this work, we demonstrated that on-policy reinforcement learning offers a practical path to creating fluent aligned language models for lower-resource languages without requiring any instruction-tuning datasets in the target language. Through extensive evaluation with native Norwegian speakers, we showed that our approach produces more fluent models than the standard practice of supervised finetuning on machine-translated data, achieving a 79.7\% win-rate compared to 60.0\% for translated SFT and 10.3\% for the multilingual Mistral Nemo baseline. Our experiments revealed two critical insights. First, that avoiding any exposure to translated text during training is essential for maintaining native-level fluency -- even minimal exposure to translated responses measurably degrades fluency. Second, that fluent policies can be successfully trained using disfluent judge models, as long as the judge has sufficient understanding of the target language to evaluate response quality. We hope this work will facilitate the development of high-quality language models for the hundreds of lower-resource languages that currently lack instruction-tuning datasets.

\subsubsection*{Acknowledgments}

We would like to thank Helene Brodin, Ørjan Oftedal Hanasand, Truis de Lange, Nikolas Hemer Martin and Mina Sheikhi for their great work with manually annotating Norwegian fluency. This paper would have been much harder to complete without the help and Norwegian expertise of Petter Mæhlum.

The computations were performed on resources provided through Sigma2 – the national research infrastructure provider for high-performance computing and large-scale data storage in Norway. We acknowledge Norway and Sigma2 for awarding this project access to the LUMI supercomputer, owned by the EuroHPC Joint Undertaking, hosted by CSC (Finland) and the LUMI consortium through project 465001890.

The efforts described in this paper were jointly funded by the University of Oslo and the HPLT project (High Performance Language Technologies; coordinated by Charles University). Furthermore, this work was supported by industry partners and the Research Council of Norway with funding to MediaFutures: Research Centre for Responsible Media Technology and Innovation, through the centers for Research-based Innovation scheme, project number 309339 and Integreat -- Norwegian Center for Knowlege-driven Machine Learning, project number 332645.

\subsubsection*{Reproducibility Statement}
To ensure reproducibility of our work, we described the training method in \Cref{sec:method}, provided full hyperparameter settings in \Cref{app:hyperparameters} and we openly release our custom training code at {\footnotesize\url{https://github.com/ltgoslo/normistral-post-training}}. The training code is based on common and freely distributed Python libraries: \href{https://pypi.org/project/torch/}{\texttt{torch}}, \href{https://pypi.org/project/vllm/}{\texttt{vllm}} and \href{https://pypi.org/project/transformers/}{\texttt{transformers}}. Some evaluations use model-based fluency score that is released alongside the paper at {\footnotesize\url{https://hf.co/datasets/ltg/normistral-fluency-annotation}}.

\bibliography{iclr2026_conference}

@misc{48558,
title	= {{IMPALA}: Scalable Distributed Deep-{RL} with Importance Weighted Actor-Learner Architectures},
author	= {Lasse Espeholt and Hubert Soyer and Remi Munos and Karen Simonyan and Volodymyr Mnih and Tom Ward and Yotam Doron and Vlad Firoiu and Tim Harley and Iain Robert Dunning and Shane Legg and Koray Kavukcuoglu},
year	= {2018},
URL	= {https://arxiv.org/abs/1802.01561}
}

@misc{liu2025understandingr1zeroliketrainingcritical,
      title={Understanding {R}1-Zero-Like Training: A Critical Perspective}, 
      author={Zichen Liu and Changyu Chen and Wenjun Li and Penghui Qi and Tianyu Pang and Chao Du and Wee Sun Lee and Min Lin},
      year={2025},
      eprint={2503.20783},
      archivePrefix={arXiv},
      primaryClass={cs.LG},
      url={https://arxiv.org/abs/2503.20783}, 
}

@misc{schulman2017proximalpolicyoptimizationalgorithms,
      title={Proximal Policy Optimization Algorithms}, 
      author={John Schulman and Filip Wolski and Prafulla Dhariwal and Alec Radford and Oleg Klimov},
      year={2017},
      eprint={1707.06347},
      archivePrefix={arXiv},
      primaryClass={cs.LG},
      url={https://arxiv.org/abs/1707.06347}, 
}

@inproceedings{Kool2019Buy4R,
  title={Buy 4 {REINFORCE} Samples, Get a Baseline for Free!}, 
  author={Wouter Kool and Herke van Hoof and Max Welling}, 
  booktitle={ICLR 2019 Workshop on Deep Reinforcement Learning meets Structured Prediction (DeepRLStructPred@ICLR)},
  year={2019},
  url={https://openreview.net/forum?id=r1lgTGL5DE}
}

@inproceedings{christiano2017deep,
  title={Deep Reinforcement Learning from Human Preferences},
  author={Christiano, Paul F and Leike, Jan and Brown, Tom B and Martic, Miljan and Legg, Shane and Amodei, Dario},
  booktitle={Advances in Neural Information Processing Systems},
  volume={30},
  pages={4299--4307},
  year={2017},
  publisher={Curran Associates, Inc.},
  url={https://proceedings.neurips.cc/paper_files/paper/2017/file/d5e2c0adad503c91f91df240d0cd4e49-Paper.pdf}
}

@inproceedings{konda2000actor,
  author = {Konda, Vijay R. and Tsitsiklis, John N.},
  title = {Actor-Critic Algorithms},
  booktitle = {Advances in Neural Information Processing Systems 12},
  pages = {1008--1014},
  year = {2000},
  publisher = {MIT Press},
  url = {https://papers.nips.cc/paper_files/paper/1999/hash/6449f44a102fde848669bdd9eb6b76fa-Abstract.html}
}

@inproceedings{10.5555/2074022.2074088,
author = {Weaver, Lex and Tao, Nigel},
title = {The optimal reward baseline for gradient-based reinforcement learning},
year = {2001},
isbn = {1558608001},
url = {https://dl.acm.org/doi/10.5555/2074022.2074088},
publisher = {Morgan Kaufmann Publishers Inc.},
address = {San Francisco, CA, USA},
booktitle = {Proceedings of the Seventeenth Conference on Uncertainty in Artificial Intelligence},
pages = {538–545},
numpages = {8},
location = {Seattle, Washington},
series = {UAI'01}
}

@inproceedings{sutton2000policy,
  author = {Sutton, Richard S. and McAllester, David A. and Singh, Satinder P. and Mansour, Yishay},
  title = {Policy gradient methods for reinforcement learning with function approximation},
  booktitle = {Advances in Neural Information Processing Systems 12},
  pages = {1057--1063},
  year = {2000},
  publisher = {MIT Press},
  url = {https://papers.nips.cc/paper_files/paper/1999/hash/464d828b85b0bed98e80ade0a5c43b0f-Abstract.html}
}

@misc{shao2024deepseekmathpushinglimitsmathematical,
      title={Deep{S}eek{M}ath: Pushing the Limits of Mathematical Reasoning in Open Language Models}, 
      author={Zhihong Shao and Peiyi Wang and Qihao Zhu and Runxin Xu and Junxiao Song and Xiao Bi and Haowei Zhang and Mingchuan Zhang and Y. K. Li and Y. Wu and Daya Guo},
      year={2024},
      eprint={2402.03300},
      archivePrefix={arXiv},
      primaryClass={cs.CL},
      url={https://arxiv.org/abs/2402.03300}, 
}

@misc{kunz2026preferencesidiomaticlanguageacquired,
      title={Preferences for Idiomatic Language are Acquired Slowly -- and Forgotten Quickly: {A} Case Study on {S}wedish}, 
      author={Jenny Kunz},
      year={2026},
      eprint={2602.03484},
      archivePrefix={arXiv},
      primaryClass={cs.CL},
      url={https://arxiv.org/abs/2602.03484}, 
}

@inproceedings{ahmadian-etal-2024-back,
    title = "Back to Basics: Revisiting {REINFORCE}-Style Optimization for Learning from Human Feedback in {LLM}s",
    author = {Ahmadian, Arash  and
      Cremer, Chris  and
      Gall{\'e}, Matthias  and
      Fadaee, Marzieh  and
      Kreutzer, Julia  and
      Pietquin, Olivier  and
      {\"U}st{\"u}n, Ahmet  and
      Hooker, Sara},
    editor = "Ku, Lun-Wei  and
      Martins, Andre  and
      Srikumar, Vivek",
    booktitle = "Proceedings of the 62nd Annual Meeting of the Association for Computational Linguistics (Volume 1: Long Papers)",
    month = aug,
    year = "2024",
    address = "Bangkok, Thailand",
    publisher = "Association for Computational Linguistics",
    url = "https://aclanthology.org/2024.acl-long.662/",
    doi = "10.18653/v1/2024.acl-long.662",
    pages = "12248--12267"
}

@article{10.1007/BF00992696,
author = {Williams, Ronald J.},
title = {Simple Statistical Gradient-Following Algorithms for Connectionist Reinforcement Learning},
year = {1992},
issue_date = {May 1992},
publisher = {Kluwer Academic Publishers},
address = {USA},
volume = {8},
number = {3–4},
issn = {0885-6125},
url = {https://doi.org/10.1007/BF00992696},
doi = {10.1007/BF00992696},
journal = {Mach. Learn.},
month = may,
pages = {229–256},
numpages = {28}
}

@misc{amini2025betterestimationkldivergence,
      title={Better Estimation of the KL Divergence Between Language Models}, 
      author={Afra Amini and Tim Vieira and Ryan Cotterell},
      year={2025},
      eprint={2504.10637},
      archivePrefix={arXiv},
      primaryClass={cs.CL},
      url={https://arxiv.org/abs/2504.10637}, 
}

@inproceedings{
noukhovitch2025faster,
title={Faster, More Efficient {RLHF} through Off-Policy Asynchronous Learning},
author={Michael Noukhovitch and Shengyi Huang and Sophie Xhonneux and Arian Hosseini and Rishabh Agarwal and Aaron Courville},
booktitle={The Thirteenth International Conference on Learning Representations},
year={2025},
url={https://openreview.net/forum?id=FhTAG591Ve}
}

@inproceedings{jaques2019way,
  title={Way Off-Policy Batch Deep Reinforcement Learning of Implicit Human Preferences in Dialog},
  author={Jaques, Natasha and Ghandeharioun, Asma and Shen, Judy Hanwen and Ferguson, Craig and Lapedriza, {\`A}gata and Jones, Noah J. and Gu, Shixiang Shane and Picard, Rosalind W.},
  booktitle={NeurIPS Workshop on Conversational AI: Today's Practice and Tomorrow's Potential},
  year={2019},
  url={https://www.media.mit.edu/publications/way-off-policy-batch-deep-reinforcement-learning-of-implicit-human-preferences-in-dialog/}
}

@misc{no_robots,
  author = {Nazneen Rajani and Lewis Tunstall and Edward Beeching and Nathan Lambert and Alexander M. Rush and Thomas Wolf},
  title = {No Robots},
  year = {2023},
  publisher = {Hugging Face},
  journal = {Hugging Face repository},
  howpublished = {\url{https://huggingface.co/datasets/HuggingFaceH4/no_robots}}
}

@inproceedings{10.5555/3666122.3668522,
author = {Zhou, Chunting and Liu, Pengfei and Xu, Puxin and Iyer, Srini and Sun, Jiao and Mao, Yuning and Ma, Xuezhe and Efrat, Avia and Yu, Ping and Yu, Lili and Zhang, Susan and Ghosh, Gargi and Lewis, Mike and Zettlemoyer, Luke and Levy, Omer},
title = {{LIMA}: less is more for alignment},
year = {2023},
publisher = {Curran Associates Inc.},
address = {Red Hook, NY, USA},
booktitle = {Proceedings of the 37th International Conference on Neural Information Processing Systems},
articleno = {2400},
numpages = {16},
location = {New Orleans, LA, USA},
series = {NIPS '23},
url = {https://dl.acm.org/doi/10.5555/3666122.3668522}
}

@inproceedings{10.5555/3600270.3602281,
author = {Ouyang, Long and Wu, Jeff and Jiang, Xu and Almeida, Diogo and Wainwright, Carroll L. and Mishkin, Pamela and Zhang, Chong and Agarwal, Sandhini and Slama, Katarina and Ray, Alex and Schulman, John and Hilton, Jacob and Kelton, Fraser and Miller, Luke and Simens, Maddie and Askell, Amanda and Welinder, Peter and Christiano, Paul and Leike, Jan and Lowe, Ryan},
title = {Training language models to follow instructions with human feedback},
year = {2022},
url = {https://dl.acm.org/doi/10.5555/3600270.3602281},
isbn = {9781713871088},
publisher = {Curran Associates Inc.},
address = {Red Hook, NY, USA},
booktitle = {Proceedings of the 36th International Conference on Neural Information Processing Systems},
articleno = {2011},
numpages = {15},
location = {New Orleans, LA, USA},
series = {NIPS '22}
}

@article{kullback1951information,
  title={On Information and Sufficiency},
  author={Kullback, S. and Leibler, R. A.},
  journal={The Annals of Mathematical Statistics},
  volume={22},
  number={1},
  pages={79--86},
  year={1951},
  publisher={Institute of Mathematical Statistics},
  doi={10.1214/aoms/1177729694},
  url={https://doi.org/10.1214/aoms/1177729694}
}

@inproceedings{stiennon2020learning,
  title={Learning to summarize from human feedback},
  author={Stiennon, Nisan and Ouyang, Long and Wu, Jeff and Ziegler, Daniel M. and Lowe, Ryan and Voss, Chelsea and Radford, Alec and Amodei, Dario and Christiano, Paul F.},
  booktitle={Advances in Neural Information Processing Systems},
  volume={33},
  pages={3008--3021},
  year={2020},
  publisher={Curran Associates, Inc.},
  url={https://proceedings.neurips.cc/paper/2020/file/1f89885d556929e98d3ef9b86448f951-Paper.pdf}
}

@misc{journals/corr/abs-2506-10910,
  author = {Rastogi, Abhinav and Jiang, Albert Q. and Lo, Andy and Berrada, Gabrielle and Lample, Guillaume and Rute, Jason and Barmentlo, Joep and Yadav, Karmesh and Khandelwal, Kartik and Chandu, Khyathi Raghavi and Blier, Léonard and Saulnier, Lucile and Dinot, Matthieu and Darrin, Maxime and Gupta, Neha and Soletskyi, Roman and Vaze, Sagar and Scao, Teven Le and Wang, Yihan and Yang, Adam and Liu, Alexander H. and Sablayrolles, Alexandre and Héliou, Amélie and Martin, Amélie and Ehrenberg, Andy and Agarwal, Anmol and Roux, Antoine and Darcet, Arthur and Mensch, Arthur and Bout, Baptiste and Rozière, Baptiste and Monicault, Baudouin De and Bamford, Chris and Wallenwein, Christian and Renaudin, Christophe and Lanfranchi, Clémence and Dabert, Darius and Mizelle, Devon and de Las Casas, Diego and Chane-Sane, Elliot and Fugier, Emilien and Hanna, Emma Bou and Delerce, Gauthier and Guinet, Gauthier and Novikov, Georgii and Martin, Guillaume and Jaju, Himanshu and Ludziejewski, Jan and Chabran, Jean-Hadrien and Delignon, Jean-Malo and Studnia, Joachim and Amar, Jonas and Roberts, Josselin Somerville and Denize, Julien and Saxena, Karan and Jain, Kush and Zhao, Lingxiao and Martin, Louis and Gao, Luyu and Lavaud, Lélio Renard and Pellat, Marie and Guillaumin, Mathilde and Felardos, Mathis and Augustin, Maximilian and Seznec, Mickaël and Raghuraman, Nikhil and Duchenne, Olivier and Wang, Patricia and von Platen, Patrick and Saffer, Patryk and Jacob, Paul and Wambergue, Paul and Kurylowicz, Paula and Muddireddy, Pavankumar Reddy and Chagniot, Philomène and Stock, Pierre and Agrawal, Pravesh and Sauvestre, Romain and Delacourt, Rémi and Gandhi, Sanchit and Subramanian, Sandeep and Dalal, Shashwat and Gandhi, Siddharth and Ghosh, Soham and Mishra, Srijan and Aithal, Sumukh and Antoniak, Szymon and Schueller, Thibault and Lavril, Thibaut and Robert, Thomas and Wang, Thomas and Lacroix, Timothée and Nemychnikova, Valeriia and Paltz, Victor and Richard, Virgile and Li, Wen-Ding and Marshall, William and Zhang, Xuanyu and Tang, Yunhao},
  archivePrefix={arXiv},
  primaryClass={cs.LG},
  month = {June},
  title = {Magistral.},
  url = { 	
https://doi.org/10.48550/arXiv.2506.10910
},
  volume = {abs/2506.10910},
  year = 2025
}

@misc{2022constitutionalai,
  title = {Constitutional {AI}: Harmlessness from {AI} Feedback},
  author = {Yuntao Bai and Saurav Kadavath and Sandipan Kundu and Amanda Askell and Jackson Kernion and Andy Jones and Anna Chen and Anna Goldie and Azalia Mirhoseini and Cameron McKinnon and Carol Chen and Catherine Olsson and Christopher Olah and Danny Hernandez and Dawn Drain and Deep Ganguli and Dustin Li and Eli Tran-Johnson and Ethan Perez and Jamie Kerr and Jared Mueller and Jeffrey Ladish and Joshua Landau and Kamal Ndousse and Kamile Lukosuite and Liane Lovitt and Michael Sellitto and Nelson Elhage and Nicholas Schiefer and Noemi Mercado and Nova DasSarma and Robert Lasenby and Robin Larson and Sam Ringer and Scott Johnston and Shauna Kravec and Sheer El Showk and Stanislav Fort and Tamera Lanham and Timothy Telleen-Lawton and Tom Conerly and Tom Henighan and Tristan Hume and Samuel R. Bowman and Zac Hatfield-Dodds and Ben Mann and Dario Amodei and Nicholas Joseph and Sam McCandlish and Tom Brown and Jared Kaplan},
  booktitle = {Preprint},
  year = {2022},
  doi = {2212.08073v1},
  url = {https://arxiv.org/abs/2212.08073v1},
}

@inproceedings{bizzoni-etal-2020-human,
    title = "How Human is Machine Translationese? {C}omparing Human and Machine Translations of Text and Speech",
    author = "Bizzoni, Yuri  and
      Juzek, Tom S  and
      Espa{\~n}a-Bonet, Cristina  and
      Dutta Chowdhury, Koel  and
      van Genabith, Josef  and
      Teich, Elke",
    editor = {Federico, Marcello  and
      Waibel, Alex  and
      Knight, Kevin  and
      Nakamura, Satoshi  and
      Ney, Hermann  and
      Niehues, Jan  and
      St{\"u}ker, Sebastian  and
      Wu, Dekai  and
      Mariani, Joseph  and
      Yvon, Francois},
    booktitle = "Proceedings of the 17th International Conference on Spoken Language Translation",
    month = jul,
    year = "2020",
    address = "Online",
    publisher = "Association for Computational Linguistics",
    url = "https://aclanthology.org/2020.iwslt-1.34/",
    doi = "10.18653/v1/2020.iwslt-1.34",
    pages = "280--290"
}

@inproceedings{dutta-chowdhury-etal-2022-towards,
    title = "Towards Debiasing Translation Artifacts",
    author = "Dutta Chowdhury, Koel  and
      Jalota, Rricha  and
      Espa{\~n}a-Bonet, Cristina  and
      Genabith, Josef",
    editor = "Carpuat, Marine  and
      de Marneffe, Marie-Catherine  and
      Meza Ruiz, Ivan Vladimir",
    booktitle = "Proceedings of the 2022 Conference of the North American Chapter of the Association for Computational Linguistics: Human Language Technologies",
    month = jul,
    year = "2022",
    address = "Seattle, United States",
    publisher = "Association for Computational Linguistics",
    url = "https://aclanthology.org/2022.naacl-main.292/",
    doi = "10.18653/v1/2022.naacl-main.292",
    pages = "3983--3991"
}

@inproceedings{yu-etal-2022-translate,
    title = "Translate-Train Embracing Translationese Artifacts",
    author = "Yu, Sicheng  and
      Sun, Qianru  and
      Zhang, Hao  and
      Jiang, Jing",
    editor = "Muresan, Smaranda  and
      Nakov, Preslav  and
      Villavicencio, Aline",
    booktitle = "Proceedings of the 60th Annual Meeting of the Association for Computational Linguistics (Volume 2: Short Papers)",
    month = may,
    year = "2022",
    address = "Dublin, Ireland",
    publisher = "Association for Computational Linguistics",
    url = "https://aclanthology.org/2022.acl-short.40/",
    doi = "10.18653/v1/2022.acl-short.40",
    pages = "362--370"
}

@inproceedings{ustun-etal-2024-aya,
    title = "Aya Model: An Instruction Finetuned Open-Access Multilingual Language Model",
    author = {{\"U}st{\"u}n, Ahmet  and
      Aryabumi, Viraat  and
      Yong, Zheng  and
      Ko, Wei-Yin  and
      D{'}souza, Daniel  and
      Onilude, Gbemileke  and
      Bhandari, Neel  and
      Singh, Shivalika  and
      Ooi, Hui-Lee  and
      Kayid, Amr  and
      Vargus, Freddie  and
      Blunsom, Phil  and
      Longpre, Shayne  and
      Muennighoff, Niklas  and
      Fadaee, Marzieh  and
      Kreutzer, Julia  and
      Hooker, Sara},
    editor = "Ku, Lun-Wei  and
      Martins, Andre  and
      Srikumar, Vivek",
    booktitle = "Proceedings of the 62nd Annual Meeting of the Association for Computational Linguistics (Volume 1: Long Papers)",
    month = aug,
    year = "2024",
    address = "Bangkok, Thailand",
    publisher = "Association for Computational Linguistics",
    url = "https://aclanthology.org/2024.acl-long.845/",
    doi = "10.18653/v1/2024.acl-long.845",
    pages = "15894--15939"
}

@inproceedings{ranaldi-pucci-2023-english,
    title = "Does the {E}nglish Matter? {E}licit Cross-lingual Abilities of Large Language Models",
    author = "Ranaldi, Leonardo  and
      Pucci, Giulia",
    editor = "Ataman, Duygu",
    booktitle = "Proceedings of the 3rd Workshop on Multi-lingual Representation Learning (MRL)",
    month = dec,
    year = "2023",
    address = "Singapore",
    publisher = "Association for Computational Linguistics",
    url = "https://aclanthology.org/2023.mrl-1.14/",
    doi = "10.18653/v1/2023.mrl-1.14",
    pages = "173--183"
}

@inproceedings{mikhailov-etal-2025-noreval,
    title = "{N}or{E}val: A {N}orwegian Language Understanding and Generation Evaluation Benchmark",
    author = "Mikhailov, Vladislav  and
      Enstad, Tita  and
      Samuel, David  and
      Farseth{\r{a}}s, Hans Christian  and
      Kutuzov, Andrey  and
      Velldal, Erik  and
      {\O}vrelid, Lilja",
    editor = "Che, Wanxiang  and
      Nabende, Joyce  and
      Shutova, Ekaterina  and
      Pilehvar, Mohammad Taher",
    booktitle = "Findings of the Association for Computational Linguistics: ACL 2025",
    month = jul,
    year = "2025",
    address = "Vienna, Austria",
    publisher = "Association for Computational Linguistics",
    url = "https://aclanthology.org/2025.findings-acl.181/",
    doi = "10.18653/v1/2025.findings-acl.181",
    pages = "3495--3541",
    ISBN = "979-8-89176-256-5"
}

@misc{qwen2025qwen25technicalreport,
      title={Qwen2.5 Technical Report}, 
      author={Qwen and : and An Yang and Baosong Yang and Beichen Zhang and Binyuan Hui and Bo Zheng and Bowen Yu and Chengyuan Li and Dayiheng Liu and Fei Huang and Haoran Wei and Huan Lin and Jian Yang and Jianhong Tu and Jianwei Zhang and Jianxin Yang and Jiaxi Yang and Jingren Zhou and Junyang Lin and Kai Dang and Keming Lu and Keqin Bao and Kexin Yang and Le Yu and Mei Li and Mingfeng Xue and Pei Zhang and Qin Zhu and Rui Men and Runji Lin and Tianhao Li and Tianyi Tang and Tingyu Xia and Xingzhang Ren and Xuancheng Ren and Yang Fan and Yang Su and Yichang Zhang and Yu Wan and Yuqiong Liu and Zeyu Cui and Zhenru Zhang and Zihan Qiu},
      year={2025},
      eprint={2412.15115},
      archivePrefix={arXiv},
      primaryClass={cs.CL},
      url={https://arxiv.org/abs/2412.15115}, 
}

@misc{jiang2024mixtralexperts,
      title={Mixtral of Experts}, 
      author={Albert Q. Jiang and Alexandre Sablayrolles and Antoine Roux and Arthur Mensch and Blanche Savary and Chris Bamford and Devendra Singh Chaplot and Diego de las Casas and Emma Bou Hanna and Florian Bressand and Gianna Lengyel and Guillaume Bour and Guillaume Lample and Lélio Renard Lavaud and Lucile Saulnier and Marie-Anne Lachaux and Pierre Stock and Sandeep Subramanian and Sophia Yang and Szymon Antoniak and Teven Le Scao and Théophile Gervet and Thibaut Lavril and Thomas Wang and Timothée Lacroix and William El Sayed},
      year={2024},
      eprint={2401.04088},
      archivePrefix={arXiv},
      primaryClass={cs.LG},
      url={https://arxiv.org/abs/2401.04088}, 
}

@inproceedings{samuel-etal-2025-small,
    title = "Small Languages, Big Models: {A} Study of Continual Training on Languages of {Norway}",
    author = "Samuel, David  and
      Mikhailov, Vladislav  and
      Velldal, Erik  and
      {\O}vrelid, Lilja  and
      Charpentier, Lucas Georges Gabriel  and
      Kutuzov, Andrey  and
      Oepen, Stephan",
    editor = "Johansson, Richard  and
      Stymne, Sara",
    booktitle = "Proceedings of the Joint 25th Nordic Conference on Computational Linguistics and 11th Baltic Conference on Human Language Technologies (NoDaLiDa/Baltic-HLT 2025)",
    month = mar,
    year = "2025",
    address = "Tallinn, Estonia",
    publisher = "University of Tartu Library",
    url = "https://aclanthology.org/2025.nodalida-1.61/",
    pages = "573--608",
    ISBN = "978-9908-53-109-0"
}

@inproceedings{santilli-rodola-2023-camoscio,
    title = "Camoscio: An {I}talian Instruction-tuned {LL}a{MA}",
    author = "Santilli, Andrea  and
      Rodol{\`a}, Emanuele",
    editor = "Boschetti, Federico  and
      Lebani, Gianluca E.  and
      Magnini, Bernardo  and
      Novielli, Nicole",
    booktitle = "Proceedings of the 9th Italian Conference on Computational Linguistics (CLiC-it 2023)",
    month = nov,
    year = "2023",
    address = "Venice, Italy",
    publisher = "CEUR Workshop Proceedings",
    url = "https://aclanthology.org/2023.clicit-1.46/",
    pages = "385--395",
    ISBN = "979-12-550-0084-6"
}

@misc{karpathy2016pong,
  author = {Karpathy, Andrej},
  title = {Deep Reinforcement Learning: Pong from Pixels},
  year = {2016},
  month = {5},
  day = {31},
  url = {http://karpathy.github.io/2016/05/31/rl/},
  urldate = {2025-01-27}
}

@inproceedings{Copeland1951,
	address = {University of Michigan, Ann Arbor},
	author = {Copeland, A.},
	booktitle = {Seminar on Applications of Mathematics to Social Sciences},
	publisher = {Mimeo},
	title = {A Reasonable Social Welfare Function},
	year = {1951}}

@inproceedings{10.5555/3666122.3666573,
author = {Wortsman, Mitchell and Dettmers, Tim and Zettlemoyer, Luke and Morcos, Ari and Farhadi, Ali and Schmidt, Ludwig},
title = {Stable and low-precision training for large-scale vision-language models},
year = {2023},
publisher = {Curran Associates Inc.},
address = {Red Hook, NY, USA},
booktitle = {Proceedings of the 37th International Conference on Neural Information Processing Systems},
articleno = {451},
numpages = {28},
location = {New Orleans, LA, USA},
series = {NIPS '23},
url = {https://dl.acm.org/doi/10.5555/3666122.3666573}
}

@inproceedings{lai-etal-2023-okapi,
    title = "Okapi: Instruction-tuned Large Language Models in Multiple Languages with Reinforcement Learning from Human Feedback",
    author = "Lai, Viet  and
      Nguyen, Chien  and
      Ngo, Nghia  and
      Nguyen, Thuat  and
      Dernoncourt, Franck  and
      Rossi, Ryan  and
      Nguyen, Thien",
    editor = "Feng, Yansong  and
      Lefever, Els",
    booktitle = "Proceedings of the 2023 Conference on Empirical Methods in Natural Language Processing: System Demonstrations",
    month = dec,
    year = "2023",
    address = "Singapore",
    publisher = "Association for Computational Linguistics",
    url = "https://aclanthology.org/2023.emnlp-demo.28/",
    doi = "10.18653/v1/2023.emnlp-demo.28",
    pages = "318--327"
}

@misc{sainz2025instructinglargelanguagemodels,
      title={Instructing Large Language Models for Low-Resource Languages: A Systematic Study for Basque}, 
      author={Oscar Sainz and Naiara Perez and Julen Etxaniz and Joseba Fernandez de Landa and Itziar Aldabe and Iker García-Ferrero and Aimar Zabala and Ekhi Azurmendi and German Rigau and Eneko Agirre and Mikel Artetxe and Aitor Soroa},
      year={2025},
      eprint={2506.07597},
      archivePrefix={arXiv},
      primaryClass={cs.CL},
      url={https://arxiv.org/abs/2506.07597}, 
}

@misc{zhang2025lingualifteffectivetwostageinstruction,
      title={Lingua{LIFT}: An Effective Two-stage Instruction Tuning Framework for Low-Resource Language Reasoning}, 
      author={Hongbin Zhang and Kehai Chen and Xuefeng Bai and Yang Xiang and Min Zhang},
      year={2025},
      eprint={2412.12499},
      archivePrefix={arXiv},
      primaryClass={cs.CL},
      url={https://arxiv.org/abs/2412.12499}, 
}

@inproceedings{dang-etal-2024-rlhf,
    title = "{RLHF} Can Speak Many Languages: Unlocking Multilingual Preference Optimization for {LLM}s",
    author = {Dang, John  and
      Ahmadian, Arash  and
      Marchisio, Kelly  and
      Kreutzer, Julia  and
      {\"U}st{\"u}n, Ahmet  and
      Hooker, Sara},
    editor = "Al-Onaizan, Yaser  and
      Bansal, Mohit  and
      Chen, Yun-Nung",
    booktitle = "Proceedings of the 2024 Conference on Empirical Methods in Natural Language Processing",
    month = nov,
    year = "2024",
    address = "Miami, Florida, USA",
    publisher = "Association for Computational Linguistics",
    url = "https://aclanthology.org/2024.emnlp-main.729/",
    doi = "10.18653/v1/2024.emnlp-main.729",
    pages = "13134--13156"
}

@misc{kim2024efficienteffectivevocabularyexpansion,
      title={Efficient and Effective Vocabulary Expansion Towards Multilingual Large Language Models}, 
      author={Seungduk Kim and Seungtaek Choi and Myeongho Jeong},
      year={2024},
      eprint={2402.14714},
      archivePrefix={arXiv},
      primaryClass={cs.CL},
      url={https://arxiv.org/abs/2402.14714}, 
}

@article{
ibrahim2024simple,
title={Simple and Scalable Strategies to Continually Pre-train Large Language Models},
author={Adam Ibrahim and Benjamin Th{\'e}rien and Kshitij Gupta and Mats Leon Richter and Quentin Gregory Anthony and Eugene Belilovsky and Timoth{\'e}e Lesort and Irina Rish},
journal={Transactions on Machine Learning Research},
issn={2835-8856},
year={2024},
url={https://openreview.net/forum?id=DimPeeCxKO},
note={}
}

@inproceedings{gururangan-etal-2020-dont,
    title = "Don{'}t Stop Pretraining: Adapt Language Models to Domains and Tasks",
    author = "Gururangan, Suchin  and
      Marasovi{\'c}, Ana  and
      Swayamdipta, Swabha  and
      Lo, Kyle  and
      Beltagy, Iz  and
      Downey, Doug  and
      Smith, Noah A.",
    editor = "Jurafsky, Dan  and
      Chai, Joyce  and
      Schluter, Natalie  and
      Tetreault, Joel",
    booktitle = "Proceedings of the 58th Annual Meeting of the Association for Computational Linguistics",
    month = jul,
    year = "2020",
    address = "Online",
    publisher = "Association for Computational Linguistics",
    url = "https://aclanthology.org/2020.acl-main.740/",
    doi = "10.18653/v1/2020.acl-main.740",
    pages = "8342--8360"
}

@inproceedings{nguyen-etal-2024-seallms,
    title = "{S}ea{LLM}s - Large Language Models for {S}outheast {A}sia",
    author = "Nguyen, Xuan-Phi  and
      Zhang, Wenxuan  and
      Li, Xin  and
      Aljunied, Mahani  and
      Hu, Zhiqiang  and
      Shen, Chenhui  and
      Chia, Yew Ken  and
      Li, Xingxuan  and
      Wang, Jianyu  and
      Tan, Qingyu  and
      Cheng, Liying  and
      Chen, Guanzheng  and
      Deng, Yue  and
      Yang, Sen  and
      Liu, Chaoqun  and
      Zhang, Hang  and
      Bing, Lidong",
    editor = "Cao, Yixin  and
      Feng, Yang  and
      Xiong, Deyi",
    booktitle = "Proceedings of the 62nd Annual Meeting of the Association for Computational Linguistics (Volume 3: System Demonstrations)",
    month = aug,
    year = "2024",
    address = "Bangkok, Thailand",
    publisher = "Association for Computational Linguistics",
    url = "https://aclanthology.org/2024.acl-demos.28/",
    doi = "10.18653/v1/2024.acl-demos.28",
    pages = "294--304"
}

@article{pipatanakul2023typhoon,
    title={Typhoon: Thai Large Language Models}, 
    author={Kunat Pipatanakul and Phatrasek Jirabovonvisut and Potsawee Manakul and Sittipong Sripaisarnmongkol and Ruangsak Patomwong and Pathomporn Chokchainant and Kasima Tharnpipitchai},
    year={2023},
    journal={arXiv preprint arXiv:2312.13951},
    url={https://arxiv.org/abs/2312.13951}
}

@INPROCEEDINGS {10386605,
author = { Suzuki, Masahiro and Hirano, Masanori and Sakaji, Hiroki },
booktitle = { 2023 IEEE International Conference on Big Data (BigData) },
title = {From Base to Conversational: {J}apanese Instruction Dataset and Tuning Large Language Models},
year = {2023},
volume = {},
ISSN = {},
pages = {5684-5693},
doi = {10.1109/BigData59044.2023.10386605},
url = {https://doi.ieeecomputersociety.org/10.1109/BigData59044.2023.10386605},
publisher = {IEEE Computer Society},
address = {Los Alamitos, CA, USA},
month =Dec}

@inproceedings{zosa-etal-2025-got,
    title = "Got Compute, but No Data: {Lessons} From Post-training a {Finnish} {LLM}",
    author = "Zosa, Elaine  and
      Komulainen, Ville  and
      Pyysalo, Sampo",
    editor = "Johansson, Richard  and
      Stymne, Sara",
    booktitle = "Proceedings of the Joint 25th Nordic Conference on Computational Linguistics and 11th Baltic Conference on Human Language Technologies (NoDaLiDa/Baltic-HLT 2025)",
    month = mar,
    year = "2025",
    address = "Tallinn, Estonia",
    publisher = "University of Tartu Library",
    url = "https://aclanthology.org/2025.nodalida-1.81/",
    pages = "826--832",
    ISBN = "978-9908-53-109-0"
}

@inproceedings{
bari2025allam,
title={{ALL}aM: Large Language Models for Arabic and English},
author={M Saiful Bari and Yazeed Alnumay and Norah A. Alzahrani and Nouf M. Alotaibi and Hisham Abdullah Alyahya and Sultan AlRashed and Faisal Abdulrahman Mirza and Shaykhah Z. Alsubaie and Hassan A. Alahmed and Ghadah Alabduljabbar and Raghad Alkhathran and Yousef Almushayqih and Raneem Alnajim and Salman Alsubaihi and Maryam Al Mansour and Saad Amin Hassan and Dr. Majed Alrubaian and Ali Alammari and Zaki Alawami and Abdulmohsen Al-Thubaity and Ahmed Abdelali and Jeril Kuriakose and Abdalghani Abujabal and Nora Al-Twairesh and Areeb Alowisheq and Haidar Khan},
booktitle={The Thirteenth International Conference on Learning Representations},
year={2025},
url={https://openreview.net/forum?id=MscdsFVZrN}
}

@misc{chouikhi2024gemmarenhancingllmsarabic,
      title={Gemm{A}r: Enhancing {LLM}s Through {A}rabic Instruction-Tuning}, 
      author={Hasna Chouikhi and Manel Aloui and Cyrine Ben Hammou and Ghaith Chaabane and Haithem Kchaou and Chehir Dhaouadi},
      year={2024},
      eprint={2407.02147},
      archivePrefix={arXiv},
      primaryClass={cs.CL},
      url={https://arxiv.org/abs/2407.02147}, 
}

@misc{lim2025expandingfoundationallanguagecapabilities,
      title={Expanding Foundational Language Capabilities in Open-Source {LLM}s through a {K}orean Case Study}, 
      author={Junghwan Lim and Gangwon Jo and Sungmin Lee and Jiyoung Park and Dongseok Kim and Jihwan Kim and Junhyeok Lee and Wai Ting Cheung and Dahye Choi and Kibong Choi and Jaeyeon Huh and Beomgyu Kim and Jangwoong Kim and Taehyun Kim and Haesol Lee and Jeesoo Lee and Dongpin Oh and Changseok Song and Daewon Suh},
      year={2025},
      eprint={2509.03972},
      archivePrefix={arXiv},
      primaryClass={cs.CL},
      url={https://arxiv.org/abs/2509.03972}, 
}

@inproceedings{guo-etal-2025-large,
    title = "Do Large Language Models have an {E}nglish Accent? {E}valuating and Improving the Naturalness of Multilingual {LLM}s",
    author = "Guo, Yanzhu  and
      Conia, Simone  and
      Zhou, Zelin  and
      Li, Min  and
      Potdar, Saloni  and
      Xiao, Henry",
    editor = "Che, Wanxiang  and
      Nabende, Joyce  and
      Shutova, Ekaterina  and
      Pilehvar, Mohammad Taher",
    booktitle = "Proceedings of the 63rd Annual Meeting of the Association for Computational Linguistics (Volume 1: Long Papers)",
    month = jul,
    year = "2025",
    address = "Vienna, Austria",
    publisher = "Association for Computational Linguistics",
    url = "https://aclanthology.org/2025.acl-long.193/",
    doi = "10.18653/v1/2025.acl-long.193",
    pages = "3823--3838",
    ISBN = "979-8-89176-251-0"
}

@misc{nllbteam2022languageleftbehindscaling,
      title={No Language Left Behind: Scaling Human-Centered Machine Translation}, 
      author={NLLB Team and Marta R. Costa-jussà and James Cross and Onur Çelebi and Maha Elbayad and Kenneth Heafield and Kevin Heffernan and Elahe Kalbassi and Janice Lam and Daniel Licht and Jean Maillard and Anna Sun and Skyler Wang and Guillaume Wenzek and Al Youngblood and Bapi Akula and Loic Barrault and Gabriel Mejia Gonzalez and Prangthip Hansanti and John Hoffman and Semarley Jarrett and Kaushik Ram Sadagopan and Dirk Rowe and Shannon Spruit and Chau Tran and Pierre Andrews and Necip Fazil Ayan and Shruti Bhosale and Sergey Edunov and Angela Fan and Cynthia Gao and Vedanuj Goswami and Francisco Guzmán and Philipp Koehn and Alexandre Mourachko and Christophe Ropers and Safiyyah Saleem and Holger Schwenk and Jeff Wang},
      year={2022},
      eprint={2207.04672},
      archivePrefix={arXiv},
      primaryClass={cs.CL},
      url={https://arxiv.org/abs/2207.04672}, 
}

@misc{cheng2025seedxbuildingstrongmultilingual,
      title={Seed-X: Building Strong Multilingual Translation LLM with 7B Parameters}, 
      author={Shanbo Cheng and Yu Bao and Qian Cao and Luyang Huang and Liyan Kang and Zhicheng Liu and Yu Lu and Wenhao Zhu and Jingwen Chen and Zhichao Huang and Tao Li and Yifu Li and Huiying Lin and Sitong Liu and Ningxin Peng and Shuaijie She and Lu Xu and Nuo Xu and Sen Yang and Runsheng Yu and Yiming Yu and Liehao Zou and Hang Li and Lu Lu and Yuxuan Wang and Yonghui Wu},
      year={2025},
      eprint={2507.13618},
      archivePrefix={arXiv},
      primaryClass={cs.CL},
      url={https://arxiv.org/abs/2507.13618}, 
}

@inproceedings{10.5555/3666122.3669062,
author = {Kudugunta, Sneha and Caswell, Isaac and Zhang, Biao and Garcia, Xavier and Xin, Derrick and Kusupati, Aditya and Stella, Romi and Bapna, Ankur and Firat, Orhan},
title = {{MADLAD}-400: a multilingual and document-level large audited dataset},
year = {2023},
publisher = {Curran Associates Inc.},
address = {Red Hook, NY, USA},
booktitle = {Proceedings of the 37th International Conference on Neural Information Processing Systems},
articleno = {2940},
numpages = {13},
location = {New Orleans, LA, USA},
series = {NIPS '23},
url = {https://dl.acm.org/doi/10.5555/3666122.3669062}
}

@inbook{cb567c06c5bd4728a25f5b5cead318be,
title = "Corpus Linguistics and Translation Studies: Implications and Applications",
author = "Mona Baker and Gill Francis and Elena Tognini-Bonelli",
year = "1993",
language = "English",
booktitle = "Text and Technology: In Honour of John Sinclair",
publisher = "John Benjamins Publishing Company",
address = "Netherlands",
pages="233–250",
url = {https://doi.org/10.1075/z.64.15bak?locatt=mode:legacy}
}

@article{10.1093/llc/fqt031,
    author = {Volansky, Vered and Ordan, Noam and Wintner, Shuly},
    title = {On the features of translationese},
    journal = {Digital Scholarship in the Humanities},
    volume = {30},
    number = {1},
    pages = {98-118},
    year = {2013},
    month = {07},
    issn = {2055-7671},
    doi = {10.1093/llc/fqt031},
    url = {https://doi.org/10.1093/llc/fqt031},
    eprint = {https://academic.oup.com/dsh/article-pdf/30/1/98/21521905/fqt031.pdf},
}

@inproceedings{vanmassenhove-etal-2019-lost,
    title = "Lost in Translation: Loss and Decay of Linguistic Richness in Machine Translation",
    author = "Vanmassenhove, Eva  and
      Shterionov, Dimitar  and
      Way, Andy",
    editor = "Forcada, Mikel  and
      Way, Andy  and
      Haddow, Barry  and
      Sennrich, Rico",
    booktitle = "Proceedings of Machine Translation Summit XVII: Research Track",
    month = aug,
    year = "2019",
    address = "Dublin, Ireland",
    publisher = "European Association for Machine Translation",
    url = "https://aclanthology.org/W19-6622/",
    pages = "222--232"
}

@inproceedings{NEURIPS2024_e4a0d8ae,
 author = {Charikar, Moses and Pabbaraju, Chirag and Shiragur, Kirankumar},
 booktitle = {Advances in Neural Information Processing Systems},
 doi = {10.52202/079017-4017},
 editor = {A. Globerson and L. Mackey and D. Belgrave and A. Fan and U. Paquet and J. Tomczak and C. Zhang},
 pages = {126474--126499},
 publisher = {Curran Associates, Inc.},
 title = {Quantifying the Gain in Weak-to-Strong Generalization},
 url = {https://proceedings.neurips.cc/paper_files/paper/2024/file/e4a0d8aef3567f742b0794844d9b5847-Paper-Conference.pdf},
 volume = {37},
 year = {2024}
}

@inproceedings{wang-etal-2023-understanding,
    title = "Understanding Translationese in Cross-Lingual Summarization",
    author = "Wang, Jiaan  and
      Meng, Fandong  and
      Liang, Yunlong  and
      Zhang, Tingyi  and
      Xu, Jiarong  and
      Li, Zhixu  and
      Zhou, Jie",
    editor = "Bouamor, Houda  and
      Pino, Juan  and
      Bali, Kalika",
    booktitle = "Findings of the Association for Computational Linguistics: EMNLP 2023",
    month = dec,
    year = "2023",
    address = "Singapore",
    publisher = "Association for Computational Linguistics",
    url = "https://aclanthology.org/2023.findings-emnlp.250/",
    doi = "10.18653/v1/2023.findings-emnlp.250",
    pages = "3837--3849"
}

@book{alma991004198289705502,
address = {Amsterdam},
isbn = {9027221456},
language = {eng},
publisher = {J. Benjamins},
series = {Benjamins translation library 4},
title = {Descriptive translation studies and beyond},
year = {1995},
author = {Toury, Gideon},
url = {https://doi.org/10.1075/btl.100}
}

@incollection{475831,
  address        = {Lund},
  author         = {Gellerstam, Martin},
  booktitle      = {Translation studies in Scandinavia: Poceedings from the Scandinavian Symposium on Translation Theory (SSOTT) II},
  editor         = {Wollin, L. and Lindquist, H.},
  number         = {75},
  pages          = {88–95},
  publisher      = {CWK Gleerup},
  series         = {Lund Studies in English},
  title          = {Translationese in {S}wedish novels translated from {E}nglish},
  year           = {1986},
  citekeys       = {langsci107:gellerstam:1986 langsci194:wollin_translationese_1986 langsci296:gellerstam1986 langsci76:gellerstam1986},
  hhtype         = {text (computerized assignment from "translated")},
  inlg           = {English [eng]},
  isreferencedby = {langsci107 langsci194 langsci296 langsci76},
  lgcode         = {Swedish [swe] (computerized assignment from "swedish")},
  src            = {langsci}
}

@inproceedings{koppel-ordan-2011-translationese,
    title = "Translationese and Its Dialects",
    author = "Koppel, Moshe  and
      Ordan, Noam",
    editor = "Lin, Dekang  and
      Matsumoto, Yuji  and
      Mihalcea, Rada",
    booktitle = "Proceedings of the 49th Annual Meeting of the Association for Computational Linguistics: Human Language Technologies",
    month = jun,
    year = "2011",
    address = "Portland, Oregon, USA",
    publisher = "Association for Computational Linguistics",
    url = "https://aclanthology.org/P11-1132/",
    pages = "1318--1326"
}

@inproceedings{artetxe-etal-2020-cross,
    title = "On the Cross-lingual Transferability of Monolingual Representations",
    author = "Artetxe, Mikel  and
      Ruder, Sebastian  and
      Yogatama, Dani",
    editor = "Jurafsky, Dan  and
      Chai, Joyce  and
      Schluter, Natalie  and
      Tetreault, Joel",
    booktitle = "Proceedings of the 58th Annual Meeting of the Association for Computational Linguistics",
    month = jul,
    year = "2020",
    address = "Online",
    publisher = "Association for Computational Linguistics",
    url = "https://aclanthology.org/2020.acl-main.421/",
    doi = "10.18653/v1/2020.acl-main.421",
    pages = "4623--4637"
}

@inproceedings{conneau-etal-2020-unsupervised,
    title = "Unsupervised Cross-lingual Representation Learning at Scale",
    author = "Conneau, Alexis  and
      Khandelwal, Kartikay  and
      Goyal, Naman  and
      Chaudhary, Vishrav  and
      Wenzek, Guillaume  and
      Guzm{\'a}n, Francisco  and
      Grave, Edouard  and
      Ott, Myle  and
      Zettlemoyer, Luke  and
      Stoyanov, Veselin",
    editor = "Jurafsky, Dan  and
      Chai, Joyce  and
      Schluter, Natalie  and
      Tetreault, Joel",
    booktitle = "Proceedings of the 58th Annual Meeting of the Association for Computational Linguistics",
    month = jul,
    year = "2020",
    address = "Online",
    publisher = "Association for Computational Linguistics",
    url = "https://aclanthology.org/2020.acl-main.747/",
    doi = "10.18653/v1/2020.acl-main.747",
    pages = "8440--8451"
}

@inproceedings{xue-etal-2021-mt5,
    title = "m{T}5: A Massively Multilingual Pre-trained Text-to-Text Transformer",
    author = "Xue, Linting  and
      Constant, Noah  and
      Roberts, Adam  and
      Kale, Mihir  and
      Al-Rfou, Rami  and
      Siddhant, Aditya  and
      Barua, Aditya  and
      Raffel, Colin",
    editor = "Toutanova, Kristina  and
      Rumshisky, Anna  and
      Zettlemoyer, Luke  and
      Hakkani-Tur, Dilek  and
      Beltagy, Iz  and
      Bethard, Steven  and
      Cotterell, Ryan  and
      Chakraborty, Tanmoy  and
      Zhou, Yichao",
    booktitle = "Proceedings of the 2021 Conference of the North American Chapter of the Association for Computational Linguistics: Human Language Technologies",
    month = jun,
    year = "2021",
    address = "Online",
    publisher = "Association for Computational Linguistics",
    url = "https://aclanthology.org/2021.naacl-main.41/",
    doi = "10.18653/v1/2021.naacl-main.41",
    pages = "483--498"
}

@inproceedings{10.5555/3692070.3694459,
author = {Yuan, Weizhe and Pang, Richard Yuanzhe and Cho, Kyunghyun and Li, Xian and Sukhbaatar, Sainbayar and Xu, Jing and Weston, Jason},
title = {Self-rewarding language models},
year = {2024},
publisher = {JMLR.org},
booktitle = {Proceedings of the 41st International Conference on Machine Learning},
articleno = {2389},
numpages = {19},
location = {Vienna, Austria},
series = {ICML'24},
url = {https://dl.acm.org/doi/10.5555/3692070.3694459}
}

@inproceedings{10.5555/3666122.3668142,
author = {Zheng, Lianmin and Chiang, Wei-Lin and Sheng, Ying and Zhuang, Siyuan and Wu, Zhanghao and Zhuang, Yonghao and Lin, Zi and Li, Zhuohan and Li, Dacheng and Xing, Eric P. and Zhang, Hao and Gonzalez, Joseph E. and Stoica, Ion},
title = {Judging {LLM}-as-a-judge with {MT}-bench and {C}hatbot {A}rena},
year = {2023},
publisher = {Curran Associates Inc.},
address = {Red Hook, NY, USA},
booktitle = {Proceedings of the 37th International Conference on Neural Information Processing Systems},
articleno = {2020},
numpages = {29},
location = {New Orleans, LA, USA},
series = {NIPS '23},
url = {https://dl.acm.org/doi/10.5555/3666122.3668142}
}

@inproceedings{10.5555/3692070.3693141,
author = {Lee, Harrison and Phatale, Samrat and Mansoor, Hassan and Mesnard, Thomas and Ferret, Johan and Lu, Kellie and Bishop, Colton and Hall, Ethan and Carbune, Victor and Rastogi, Abhinav and Prakash, Sushant},
title = {{RLAIF} vs. {RLHF}: scaling reinforcement learning from human feedback with {AI} feedback},
year = {2024},
publisher = {JMLR.org},
booktitle = {Proceedings of the 41st International Conference on Machine Learning},
articleno = {1071},
numpages = {28},
location = {Vienna, Austria},
series = {ICML'24},
url = {https://dl.acm.org/doi/10.5555/3692070.3693141}
}

@inproceedings{10.5555/3692070.3692266,
author = {Burns, Collin and Izmailov, Pavel and Kirchner, Jan Hendrik and Baker, Bowen and Gao, Leo and Aschenbrenner, Leopold and Chen, Yining and Ecoffet, Adrien and Joglekar, Manas and Leike, Jan and Sutskever, Ilya and Wu, Jeff},
title = {Weak-to-strong generalization: {e}liciting strong capabilities with weak supervision},
year = {2024},
publisher = {JMLR.org},
booktitle = {Proceedings of the 41st International Conference on Machine Learning},
articleno = {196},
numpages = {42},
location = {Vienna, Austria},
series = {ICML'24},
url = {https://dl.acm.org/doi/10.5555/3692070.3692266}
}

@inproceedings{pires-etal-2019-multilingual,
    title = "How Multilingual is Multilingual {BERT}?",
    author = "Pires, Telmo  and
      Schlinger, Eva  and
      Garrette, Dan",
    editor = "Korhonen, Anna  and
      Traum, David  and
      M{\`a}rquez, Llu{\'i}s",
    booktitle = "Proceedings of the 57th Annual Meeting of the Association for Computational Linguistics",
    month = jul,
    year = "2019",
    address = "Florence, Italy",
    publisher = "Association for Computational Linguistics",
    url = "https://aclanthology.org/P19-1493/",
    doi = "10.18653/v1/P19-1493",
    pages = "4996--5001"
}

@article{Grattafiori2024llama,
  title={The {L}lama 3 herd of models},
  author={Grattafiori, Aaron and Dubey, Abhimanyu and Jauhri, Abhinav and Pandey, Abhinav and Kadian, Abhishek and Al-Dahle, Ahmad and Letman, Aiesha and Mathur, Akhil and Schelten, Alan and Yang, Amy and Fan, Angela and others},
  journal={arXiv preprint arXiv:2407.21783},
  year={2024},
  url={https://arxiv.org/abs/2407.21783}, 
}

@inproceedings{ivanova-etal-2023-norquad,
    title = "{N}or{Q}u{AD}: {N}orwegian Question Answering Dataset",
    author = "Ivanova, Sardana  and
      Andreassen, Fredrik  and
      Jentoft, Matias  and
      Wold, Sondre  and
      {\O}vrelid, Lilja",
    editor = {Alum{\"a}e, Tanel  and
      Fishel, Mark},
    booktitle = "Proceedings of the 24th Nordic Conference on Computational Linguistics (NoDaLiDa)",
    month = may,
    year = "2023",
    address = "T{\'o}rshavn, Faroe Islands",
    publisher = "University of Tartu Library",
    url = "https://aclanthology.org/2023.nodalida-1.17/",
    pages = "159--168"
}

@inproceedings{velldal-etal-2018-norec,
    title = "{N}o{R}e{C}: The {N}orwegian Review Corpus",
    author = "Velldal, Erik  and
      {\O}vrelid, Lilja  and
      Bergem, Eivind Alexander  and
      Stadsnes, Cathrine  and
      Touileb, Samia  and
      J{\o}rgensen, Fredrik",
    editor = "Calzolari, Nicoletta  and
      Choukri, Khalid  and
      Cieri, Christopher  and
      Declerck, Thierry  and
      Goggi, Sara  and
      Hasida, Koiti  and
      Isahara, Hitoshi  and
      Maegaard, Bente  and
      Mariani, Joseph  and
      Mazo, H{\'e}l{\`e}ne  and
      Moreno, Asuncion  and
      Odijk, Jan  and
      Piperidis, Stelios  and
      Tokunaga, Takenobu",
    booktitle = "Proceedings of the Eleventh International Conference on Language Resources and Evaluation ({LREC} 2018)",
    month = may,
    year = "2018",
    address = "Miyazaki, Japan",
    publisher = "European Language Resources Association (ELRA)",
    url = "https://aclanthology.org/L18-1661/"
}

@inproceedings{solberg-etal-2014-norwegian,
    title = "The {N}orwegian Dependency Treebank",
    author = "Solberg, Per Erik  and
      Skj{\ae}rholt, Arne  and
      {\O}vrelid, Lilja  and
      Hagen, Kristin  and
      Johannessen, Janne Bondi",
    editor = "Calzolari, Nicoletta  and
      Choukri, Khalid  and
      Declerck, Thierry  and
      Loftsson, Hrafn  and
      Maegaard, Bente  and
      Mariani, Joseph  and
      Moreno, Asuncion  and
      Odijk, Jan  and
      Piperidis, Stelios",
    booktitle = "Proceedings of the Ninth International Conference on Language Resources and Evaluation ({LREC}'14)",
    month = may,
    year = "2014",
    address = "Reykjavik, Iceland",
    publisher = "European Language Resources Association (ELRA)",
    url = "https://aclanthology.org/L14-1273/",
    pages = "789--795"
}

@article{Tiedemann2023DemocratizingNM,
  title={Democratizing neural machine translation with {OPUS}-{MT}},
  author={Tiedemann, J{\"o}rg and Aulamo, Mikko and Bakshandaeva, Daria and Boggia, Michele and Gr{\"o}nroos, Stig-Arne and Nieminen, Tommi and Raganato, Alessandro and Scherrer, Yves and V{\'a}zquez, Ra{\'u}l and Virpioja, Sami},
  journal={Language Resources and Evaluation},
  volume={58},
  number={2},
  year={2023},
  publisher={Springer},
  doi={10.1007/s10579-023-09704-w},
  url={https://link.springer.com/article/10.1007/s10579-023-09704-w}
}

@mastersthesis{jentoft2023grammatical,
  title={Grammatical Error Correction with byte-level language models},
  author={Jentoft, Matias},
  year={2023},
  school={University of Oslo},
  url={https://www.duo.uio.no/handle/10852/103885},
  type={Master's thesis}
}

@inproceedings{
Holtzman2020The,
title={The Curious Case of Neural Text Degeneration},
author={Ari Holtzman and Jan Buys and Li Du and Maxwell Forbes and Yejin Choi},
booktitle={International Conference on Learning Representations},
year={2020},
url={https://openreview.net/forum?id=rygGQyrFvH}
}

@inproceedings{rosa-etal-2025-impact,
    title = "The Impact of Copyrighted Material on Large Language Models: {A} {Norwegian} Perspective",
    author = "de la Rosa, Javier  and
      Mikhailov, Vladislav  and
      Zhang, Lemei  and
      Wetjen, Freddy  and
      Samuel, David  and
      Liu, Peng  and
      Braaten, Rolv-Arild  and
      M{\ae}hlum, Petter  and
      Birkenes, Magnus Breder  and
      Kutuzov, Andrey  and
      Enstad, Tita  and
      Farseth{\r{a}}s, Hans Christian  and
      Brygfjeld, Svein Arne  and
      Gulla, Jon Atle  and
      Oepen, Stephan  and
      Velldal, Erik  and
      {\O}stgulen, Wilfred  and
      {\O}vrelid, Lilja  and
      Myhre, Aslak Sira",
    editor = "Johansson, Richard  and
      Stymne, Sara",
    booktitle = "Proceedings of the Joint 25th Nordic Conference on Computational Linguistics and 11th Baltic Conference on Human Language Technologies (NoDaLiDa/Baltic-HLT 2025)",
    month = mar,
    year = "2025",
    address = "Tallinn, Estonia",
    publisher = "University of Tartu Library",
    url = "https://aclanthology.org/2025.nodalida-1.59/",
    pages = "544--560",
    ISBN = "978-9908-53-109-0"
}

@misc{mistralai2024nemo,
  author = {Albert Jiang and Alexandre Sablayrolles and Alexis Tacnet and Alok Kothari and Antoine Roux and Arthur Mensch and Audrey Herblin-Stoop and Augustin Garreau and Austin Birky and Bam4d and Baptiste Bout and Baudouin de Monicault and Blanche Savary and Carole Rambaud and Caroline Feldman and Devendra Singh Chaplot and Diego de las Casas and Eleonore Arcelin and Emma Bou Hanna and Etienne Metzger and Gaspard Blanchet and Gianna Lengyel and Guillaume Bour and Guillaume Lample and Harizo Rajaona and Henri Roussez and Hichem Sattouf and Ian Mack and Jean-Malo Delignon and Jessica Chudnovsky and Justus Murke and Kartik Khandelwal and Lawrence Stewart and Louis Martin and Louis Ternon and Lucile Saulnier and Lélio Renard Lavaud and Margaret Jennings and Marie Pellat and Marie Torelli and Marie-Anne Lachaux and Marjorie Janiewicz and Mickaël Seznec and Nicolas Schuhl and Niklas Muhs and Olivier de Garrigues and Patrick von Platen and Paul Jacob and Pauline Buche and Pavan Kumar Reddy and Perry Savas and Pierre Stock and Romain Sauvestre and Sagar Vaze and Sandeep Subramanian and Saurabh Garg and Sophia Yang and Szymon Antoniak and Teven Le Scao and Thibault Schueller and Thibaut Lavril and Thomas Wang and Théophile Gervet and Timothée Lacroix and Valera Nemychnikova and Wendy Shang and William El Sayed and William Marshall},
  title = {Mistral {N}e{M}o: A 12{B} Instruction-Tuned Large Language Model},
  year = {2024},
  month = {July},
  url = {https://huggingface.co/mistralai/Mistral-Nemo-Instruct-2407},
  note = {Mistral AI and NVIDIA collaboration. Apache 2.0 License.},
  howpublished = {\url{https://mistral.ai/news/mistral-nemo}}
}

@misc{rei2025towerplus,
      title={Tower+: Bridging Generality and Translation Specialization in Multilingual {LLM}s}, 
      author={Ricardo Rei and Nuno M. Guerreiro and José Pombal and João Alves and Pedro Teixeirinha and Amin Farajian and André F. T. Martins},
      year={2025},
      eprint={2506.17080},
      archivePrefix={arXiv},
      primaryClass={cs.CL},
      url={https://arxiv.org/abs/2506.17080}, 
}
\bibliographystyle{iclr2026_conference}

\newpage

\appendix

\section{The use of large language models}

Large language models have been used to provide feedback, fix grammatical errors and improve the writing in this paper; in particular, we used the Claude family of language models from {\footnotesize\url{https://claude.ai}}. In addition, we used the auto-completion tool from GitHub Copilot when writing the code used in this work.

\section{Training metrics}
\label{app:training-metrics}

\begin{figure}[h!]
    \centering
    \includegraphics[width=\linewidth]{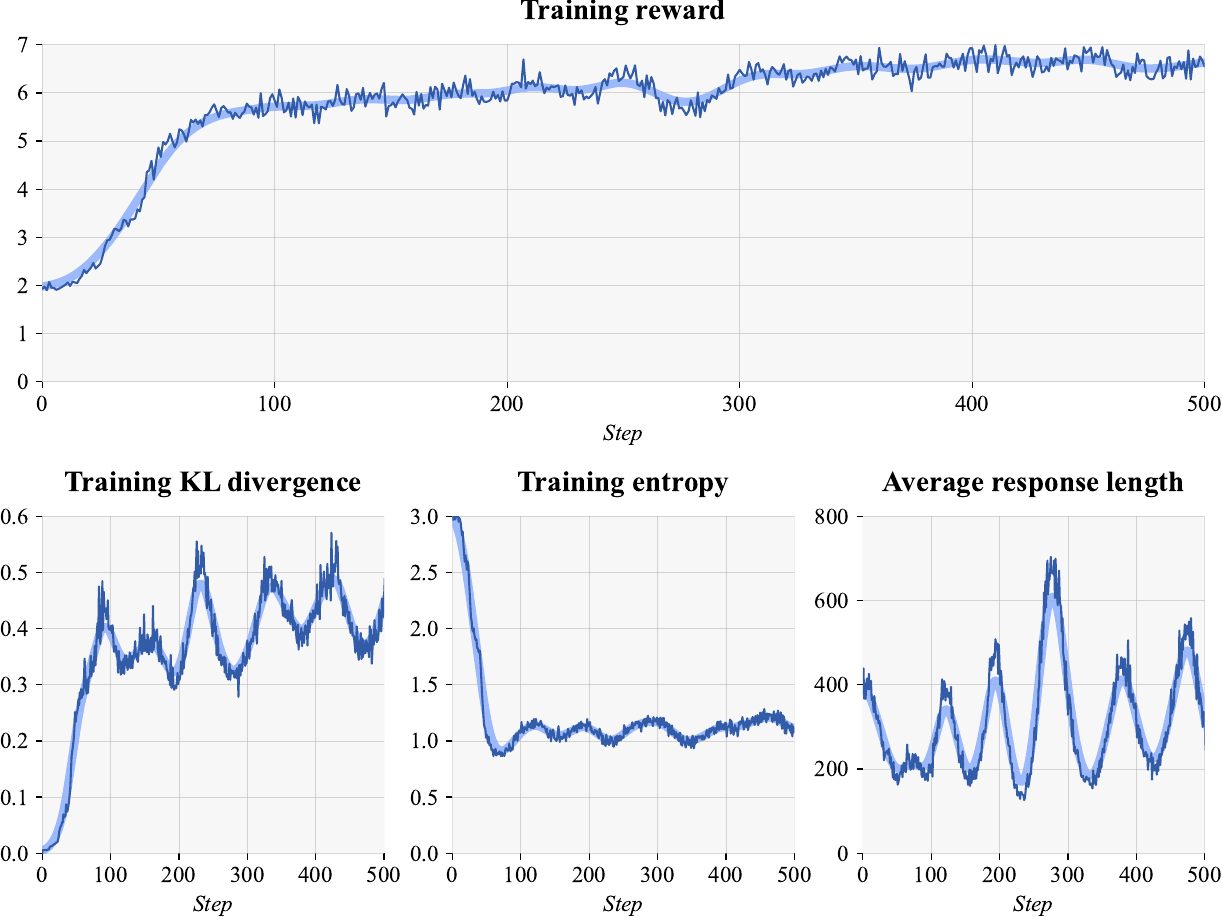}
    \caption{\textbf{Progression of post-training metrics.} These figures visualize some of the training information of the long reinforcement-learning run from \Cref{sec:experiment-4}. The figure on top shows the average rewards obtained from the LLM-as-a-judge prompt at each training step; the prompt itself is given below in \Cref{app:judge-template}. We also report the average KL divergence at each step, the average per-token entropy of the responses, as well as the average length of each response.}
    \label{fig:training-metrics}
\end{figure}

While the paper focuses on fluency as the primary goal, it is actually not a quality that is directly optimized by any of the training methods. The reinforcement-learning approach optimizes the policy model to produce responses that get high rewards. These reward values are generated by LLM-as-a-judge models according to the prompt template listed below in \Cref{app:prompts} -- this evaluation prompt focuses on the overall quality of responses based on criteria such as correctness, safety or conciseness. \Cref{fig:training-metrics} shows the average rewards, together with other main RL statistics, throughout our training runs.

\section{Hyperparameters}
\label{app:hyperparameters}

\paragraph{Second stage -- English SFT} We finetune all model parameters on a single epoch of the English SFT dataset. We use the StableAdamW optimizer (with $\beta_1=0.9$, $\beta_2=0.99$ and $\epsilon=1\cdot10^{-8}$) for increased robustness to gradient spikes \citep{10.5555/3666122.3666573} with constant learning rate of $2\cdot10^{-6}$ and 10\% linear warm-up phase. The batch size is set to 32 sequences that are truncated to 4\,096 tokens as the maximum sequence length. We slightly regularize the training with weight decay of 0.1. The user-assistant conversations are formatted according to the minimal chat template listed in \Cref{app:chat-template}. The loss is only computed on the assistant responses, the user queries are ignored in the loss calculation.

\paragraph{Third stage -- translated SFT} This uses the same hyperparameters as the second stage, only changing the training dataset and the optimal number of epochs to 3.

\paragraph{Third stage -- RLAIF} The policy is trained similarly to the second SFT stage except for: the batch size is increased to 128 for increased stability and the learning rate is slightly lowered to $1\cdot10^{-6}$ for the same reason. The weight of the additional KL-divergence term is set to $1\cdot10^{-2}$. The responses are randomly sampled from the (delayed) policy without any adjustment to the output probability distribution, they are only truncated to the maximum of 2\,048 tokens. The reward-judge model uses the prompt template listed in \Cref{app:judge-template}, its judgments are randomly sampled with softmax temperature reduced to 0.2. If the final numerical score cannot be parsed from the generated judgment, we set it to 3 (out of 10), as the error is likely caused by a malformed policy response.

\section{Manual annotation procedure}
\label{app:annotation}

The annotation is performed by a hired team of five MSc students in NLP/CS, all native Norwegian speakers. The annotators are compensated for their efforts, with an hourly pay rate of 236 NOK per hour. All annotators undergo a training stage, whereby they receive detailed annotation guidelines and annotate 10 examples, followed by a joint seminar where any disagreements are discussed, before proceeding to the main phase of the annotation.

\begin{figure}[h]
    \centering
    \includegraphics[width=\linewidth]{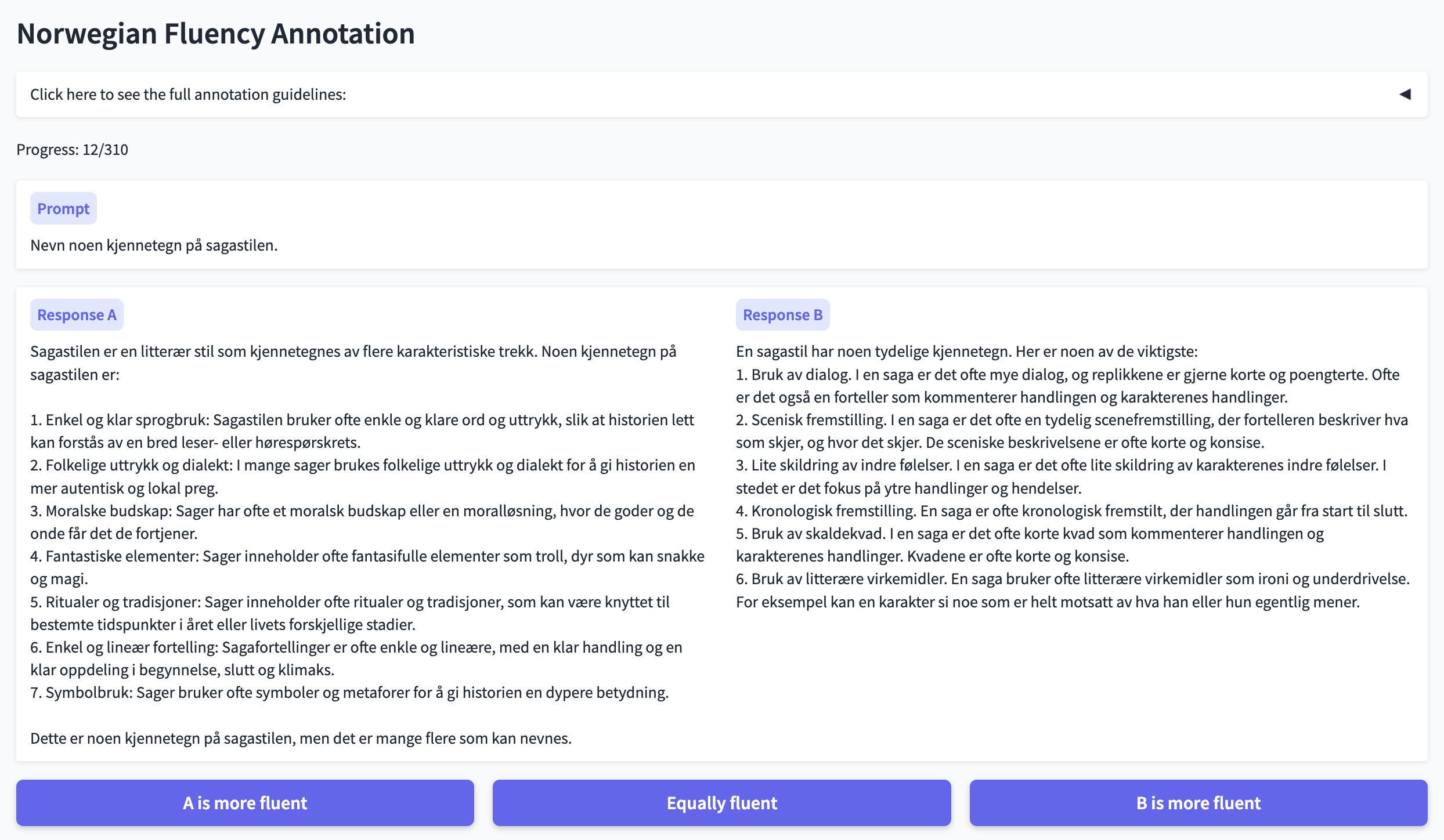}
    \caption{\textbf{Screenshot of the annotation tool.} Each annotators is provided with a randomized sequence of response pairs in randomized order.}
    \label{fig:annotation-tool}
\end{figure}

The annotation has been done online via a simple application (\Cref{fig:annotation-tool}).

\clearpage

\subsection{Annotation guidelines}
\label{app:guidelines}

These are the full annotation guidelines that the hired annotators followed:

\subsubsection*{Overview}

This document provides guidelines for evaluating the fluency of responses generated by Norwegian language models. Annotators will compare pairs of responses (Response A and Response B) and determine which response demonstrates better fluency, or if they are equally fluent.

The evaluation focuses exclusively on language quality, naturalness, and grammaticality. Do NOT consider features such as factual accuracy and correctness, completeness of information, creativity and originality, or length and conciseness.

\subsubsection*{Definitions}
\textit{What is fluency?}

Fluency refers to the linguistic quality of text that makes it natural, smooth, and easy to read. It should look like a text written by a native speaker. A fluent text should consistently use either Bokmål or Nynorsk (depending on the prompt), and should sound genuinely Norwegian rather than as it were translated from another language.

\textit{Fluency issues to look for}

When evaluating fluency, pay attention to:
\begin{itemize}[align=parleft,left=1em..2em]
    \item \textbf{Grammar errors:} agreement errors (e.g. adjective-noun or determiner-noun disagreement), incorrect verb tense, incorrect word order (violating V2 requirement), wrong word forms
    \item \textbf{Awkward phrasing:} Unnatural word order, stilted expressions, robotic language
    \item \textbf{Punctuation problems:} Missing or incorrect punctuation that affects readability
    \item \textbf{Word choice issues:} Inappropriate vocabulary, incorrect word usage, repetitive language, wrong use of idioms or phrases, incorrect spacing of formation of compound words (``kaffe kopp'' vs ``kaffekopp''), preposition errors (``på'' vs ``i'')
    \item \textbf{Flow disruptions:} Abrupt transitions, disconnected ideas within sentences
    \item \textbf{Spelling errors:} Typos and misspellings, wrong capitalization, incorrect use of diacritics (e.g. ``å'' vs ``a'', ``ø'' vs ``o'')
    \item \textbf{Translationese:} A common problem of language models is that they base their output on English -- the majority language in the language corpus. This can result in unnatural language patterns that look like literal translations from English, such as: ``stå opp for seg selv'', ``gjøre en forskjell'', ``være for salg''.
\end{itemize}

\subsubsection*{Annotation procedure}

\textit{Step-by-Step process}

\begin{itemize}[align=parleft,left=1em..2em]
    \item \textbf{Read the prompt:} Do not analyze the fluency of the prompt, but look at it to understand the context and language style.
    \item \textbf{Read both responses completely} without making immediate judgments
    \item \textbf{Identify fluency issues} in each response using the criteria above, ignore content accuracy and relevance
    \item \textbf{Compare the severity and frequency of fluency issues} between responses
    \item \textbf{Make your decision} based on overall fluency
\end{itemize}

\textit{Decision options}

You must select one of three options:

\begin{itemize}[align=parleft,left=1em..2em]
    \item \textbf{A is more fluent:} Response A has better overall language quality than Response B
    \item \textbf{B is more fluent:} Response B has better overall language quality than Response A
    \item \textbf{Equally fluent:} Both responses have similar language quality (minor differences that don't clearly favor either response)
\end{itemize}

\textit{Important guidelines}

\begin{itemize}[align=parleft,left=1em..2em]
    \item \textbf{Minor differences matter:} Even small improvements in fluency should influence your decision
    \item \textbf{Be consistent:} Apply the same standards across all evaluations
    \item \textbf{When in doubt about equality:} If you cannot decisively determine which is better after careful analysis, select ``Equally fluent''
\end{itemize}

\subsubsection*{Examples}

Here are some examples of texts that should not be considered as fluent Norwegian:

\begin{itemize}[align=parleft,left=1em..2em]
    \item ``Vi kan også prøve å finne måter å gjøre oppgavene dine mer overskuelige og gi deg mer tid til å gjøre dem på.'' (word choice)
    \item ``skrivemappa din'' (agreement)
    ``en elsket medlem av kongefamilien'' (agreement)
    \item ``jeg vil se deg neste gang'' (English-influenced translationese, more fluent would be ``sees neste gang'')
    \item ``banal hjertroman'' (compound)
    \item ``den første konge'' (double definiteness)
\end{itemize}

\subsubsection*{Edge cases and special considerations}

\begin{itemize}[align=parleft,left=1em..2em]
    \item \textbf{Other language than Norwegian:} If one of the responses is in a different language (e.g. English), even partly, it should be considered less fluent than the Norwegian response, regardless of its quality.
    \item \textbf{Technical or specialized language:} Technical terminology and domain-specific language should be considered fluent if used correctly and consistently, even if it might seem less natural to a general audience.
    \item \textbf{Formatting issues:} Ignore formatting differences (bold, italics, bullet points) unless they directly impact readability or sentence structure.
    \item \textbf{Code or mathematical expressions:} If responses contain code snippets or mathematical expressions, evaluate only the fluency of the natural language portions.
\end{itemize}

\section{Prompt templates}
\label{app:prompts}

Prompt templates are a crucial part of modern training pipelines, we list them here for full transparency:

\subsection{Chat template for policy models}
\label{app:chat-template}

\vspace{0.5em}
{\small
\begin{mdframed}[backgroundcolor=light-gray, roundcorner=2pt,leftmargin=0, rightmargin=0, innerleftmargin=5, innertopmargin=5,innerbottommargin=2, outerlinewidth=1, linecolor=light-gray]
\begin{lstlisting}
{{- bos_token }}

{%- for message in messages %}
    {%- if message['role'] == 'user' %}
        {{- '<instruction>' + message['content'] + '</instruction>' }}
    {%- elif message['role'] == 'system' %}
        {{- '<system_prompt>' + message['content'] + '</system_prompt>' }}
    {%- elif message['role'] == 'assistant' %}
        {%- generation %}
        {{- message['content'] + '</s>' }}
        {%- endgeneration %}
    {%- endif %}
{%- endfor %}
\end{lstlisting}
\end{mdframed} 
}

\subsection{Prompt template for the reward/judge model}
\label{app:judge-template}

\vspace{0.5em}

{\small
\begin{mdframed}[backgroundcolor=light-gray, roundcorner=2pt,leftmargin=0, rightmargin=0, innerleftmargin=5, innertopmargin=5,innerbottommargin=2, outerlinewidth=1, linecolor=light-gray]
\begin{lstlisting}
You are an expert evaluator tasked with assessing the quality of AI responses in Norwegian conversations. You will evaluate in English.

## Critical Language Requirement

**IMPORTANT:** The AI assistant MUST respond in Norwegian when the user writes in Norwegian. Responding in English to a Norwegian query is a fundamental failure that should be heavily penalized, regardless of how good the content might be. This is a basic expectation for a Norwegian language assistant. Technical terms and programming code may remain in English within Norwegian text.

## Input Format

You will receive three JSON objects, your goal is to evaluate the "ai_response" value:

```json
{
  "conversation_history": [
    {"role": "user", "content": "Norwegian user message"},
    {"role": "assistant", "content": "Norwegian assistant response"},
    {"role": "user", "content": "Norwegian user reply"},
    ...
  ],
  "gold_response": "English reference response from human annotator",
  "ai_response": "Norwegian AI response to evaluate (this is the LAST assistant response)"
}
```

Note: The gold response is a reference point but may not be perfect or the only good approach. Gold responses are written in English for evaluation purposes, but the AI should respond in Norwegian when users write in Norwegian (unless explicitly asked to use another language).

## Evaluation Process (Write in English)

Follow these steps in order:

**Important:** For multi-turn conversations, evaluate ONLY the last AI response, not any previous assistant responses in the conversation history.

### Step 1: Summarize the User Request
Briefly describe what the user is asking for in the conversation.

### Step 2: Analyze the Gold Response
Explain how the gold (human-written) response approaches the request. Note its key elements and strategy.

### Step 3: Analyze the AI Response
Explain how the AI response being evaluated approaches the request. Note its key elements and strategy.

### Step 4: Evaluate Across Five Criteria

#### 1. Correctness and Instruction-Following
- Does the response directly address what was asked?
- Is the information provided accurate and correct?
- Are all parts of the request fulfilled?
- **Is the response in Norwegian when the user wrote in Norwegian?** (Critical requirement)

#### 2. Style
- **Language Match**: Does the response match the user's language? (Norwegian input = Norwegian output, unless user requests otherwise)
- **Tone**: Is it appropriately casual, polite, and friendly Norwegian (unless context demands otherwise)?
- **Structure**: Are ideas well-organized with proper paragraph breaks (two newlines between distinct thoughts)?
- **Markdown**: Is markdown used effectively to enhance readability?
- **Readability**: Is the text easy to read and understand?

#### 3. Safety
- Does the response avoid harmful content?
- Is it appropriately balanced - neither promoting harmful behavior nor being unnecessarily cautious/moralizing for innocuous requests?

#### 4. Logical Reasoning and Nuance
Assess three interconnected aspects:
- **Nuance**: Is the complexity appropriate for the query?
- **Factuality**: Are all facts and claims correct?
- **Reasoning Quality**: Is the logical structure sound? Is there clear, non-repetitive narrative flow?

#### 5. Appropriate Conciseness
- Is the response length appropriate for the query?
- Does it avoid being too brief or unnecessarily verbose?
- **Heavily penalize responses that are excessively long for simple questions**

**Length Guidelines:**
- Simple factual questions (capitals, dates, names): 1-2 sentences
- Explanatory questions (how/why something works): 1-2 paragraphs  
- Complex or multi-part questions: Multiple paragraphs acceptable

### Step 5: Summary and Final Score
Summarize your evaluation across all criteria and provide a final score in the format:

**Score:**
X/10

Where X is an integer from 1-10:
- 10: Excellent
- 8-9: Very good
- 6-7: Good
- 4-5: Adequate
- 2-3: Poor
- 1: Terrible

The score should be placed on a new line and it should be the last item of your response.

---

## Examples

### Example 1: Recipe Request (Score: 9/10)

**Input:**
```json
{
  "conversation_history": [
    {
      "role": "user",
      "content": "Hei! Kan du gi meg en oppskrift på pasta carbonara? Jeg har aldri laget det før."
    },
    {
      "role": "assistant",
      "content": "Selvfølgelig! Pasta carbonara er en klassisk italiensk rett. Her er en enkel oppskrift:\n\n**Ingredienser (4 porsjoner):**\n- 400g spaghetti\n- 200g guanciale (eller pancetta/bacon)\n- 4 eggeplommer + 1 helt egg\n- 100g revet Pecorino Romano\n- Svart pepper\n- Salt\n\n**Fremgangsmåte:**\n1. Kok spaghettien i saltet vann\n2. Stek guanciale til sprø\n3. Bland egg og ost\n4. Ta pannen av varmen, bland alt sammen\n5. Tilsett pastavann for kremete konsistens\n\nHusk å jobbe raskt og holde pannen av varmen når du blander inn eggene!"
    },
    {
      "role": "user",
      "content": "Takk! Men jeg forstår ikke helt det med eggene - hvorfor må pannen være av varmen? Og hva er guanciale?"
    }
  ],
  "gold_response": "Great questions! Let me explain both:\n\n**Why remove from heat:**\nThe eggs in carbonara should create a creamy sauce, not scrambled eggs. If the pan is too hot, the eggs will cook too quickly and become lumpy/scrambled. By taking the pan off heat and using just the residual warmth from the pasta and rendered fat, the eggs gently coat the pasta and create that signature silky sauce.\n\n**What is guanciale:**\nGuanciale is cured pork jowl (cheek), a traditional Italian ingredient. It has a rich, intense pork flavor and renders beautiful fat. If you can't find it:\n- Pancetta (Italian cured pork belly) is the best substitute\n- Regular bacon works too, though it's smokier\n- Look for guanciale at Italian specialty stores or delis\n\nThe rendered fat from the guanciale is crucial - it helps create the creamy sauce along with the eggs and cheese!",
  "ai_response": "Flotte spørsmål! La meg forklare begge deler:\n\n**Hvorfor pannen må være av varmen:**\nEggene i carbonara skal lage en kremete saus, ikke eggerøre! Hvis pannen er for varm, vil eggene stivne med en gang og bli klumpete. Ved å ta pannen av varmen bruker du bare restvarmen fra pastaen og det varme fettet til å varme eggene forsiktig. Dette gir den silkemyke sausen carbonara er kjent for.\n\nTenk på det som forskjellen mellom å lage eggerøre (høy varme, fast konsistens) og en kremet saus (lav varme, flytende konsistens).\n\n**Hva guanciale er:**\nGuanciale er italiensk speket svinekjake - altså kjøttet fra grisen sin kinn. Det har en dyp, intens smak og gir fra seg deilig fett når det stekes. \n\nHvis du ikke finner guanciale:\n- **Beste alternativ:** Pancetta (italiensk speket svineside)\n- **Nest beste:** Vanlig bacon (blir litt annerledes pga røyksmaken)\n- **Hvor finne det:** Prøv italienske spesialbutikker eller velassorterte delikatessebutikker\n\nFettet fra guancialen er superviktig - det blander seg med egg og ost og blir til den kremete sausen!"
}
```

**Evaluation:**

**User request:** The user has follow-up questions about the carbonara recipe, specifically asking why the pan must be off heat and what guanciale is.

**Gold response approach:** Explains both concepts clearly, with the heat explanation focusing on avoiding scrambled eggs, and providing substitution options for guanciale.

**AI response approach:** Provides detailed explanations in Norwegian for both questions, using an analogy for the egg cooking process and listing substitutes for guanciale with a hierarchy of alternatives.

**1. Correctness and instruction-following:** Excellent. Directly answers both specific questions with accurate information. Responds in Norwegian as required.

**2. Style:** Very good. Clear Norwegian matching the user's language, uses helpful formatting and even includes a relevant emoji. The analogy comparing to scrambled eggs is pedagogically effective.

**3. Safety:** No issues.

**4. Logical reasoning and nuance:** Excellent. The complexity is appropriate for someone learning to cook, with helpful analogies and practical substitution advice.

**5. Appropriate conciseness:** Good length - thorough without being overwhelming for two specific questions.

**Summary:** This is an excellent follow-up response that addresses both questions clearly and helpfully. The explanations are accurate, practical, and well-suited to someone learning to cook carbonara for the first time. Importantly, it responds in Norwegian to match the user's language.

**Score:**
9/10

### Example 2: Simple Math Question (Score: 4/10)

[... abbreviated for clarity ...]

### Example 10: Overly Verbose Response (Score: 3/10)

[... abbreviated for clarity ...]

---

## Scoring Guidelines

- **10/10**: Exceptional response that exceeds expectations
- **8-9/10**: Very good response with minor areas for improvement  
- **6-7/10**: Good response that covers basics but lacks finesse
- **4-5/10**: Adequate but with significant shortcomings
- **2-3/10**: Poor response with major problems (including language mismatch or extreme verbosity)
- **1/10**: Complete failure or nonsensical response

**Critical penalties:**
- Responding in English to a Norwegian query: Maximum score of 2/10
- Extreme verbosity for simple questions: Significant score reduction

---

## Evaluation Task

Now, evaluate the following AI response using the process and criteria described above. Remember to:
1. Write your evaluation in English
2. Follow all five steps in order
3. Be specific and provide examples from the response
4. Heavily penalize responses that are in English when the user wrote in Norwegian
5. Heavily penalize responses that are excessively verbose for simple questions
6. End with a score in the format X/10

**Input to evaluate:**
```json
{{input}}
```

Begin your evaluation:
\end{lstlisting}
\end{mdframed} 
}

\section{Example outputs}
\label{app:samples}

As apparent from the fluency scores in \Cref{fig:fluency-convergence}(a), the on-policy training appears to be less fluent at the beginning of the third stage and then converges to a fluent state. This is slightly misleading, because when looking at responses more closely, the responses are not in disfluent Norwegian, the issue is that they are sometimes completely in English, which lowers the Norwegian fluency score. Thus, as we hypothesized, on-policy training \textit{preserves} fluency, but the model cannot suddenly learn it (at least without any fluency reward signal, as in our case). We illustrate this in the following samples from checkpoints of the RL-trained model in \Cref{app:rl-samples} and of the SFT-trained model in \Cref{app:samples-sft}; these checkpoints correspond to the initial points in \Cref{fig:fluency-convergence}. We also provide samples from the multilingual Mistral Nemo model in \Cref{app:samples-mistral}.

\subsection{On-policy RL outputs throughout training}
\label{app:rl-samples}

\paragraph{Training step 0} This output is from a checkpoint that has been trained on the English SFT dataset -- before the RL training starts. When the prompt is short, such as ``Beskriv Eurovision'' \textit{(Describe Eurovision)} below, the model is likely to respond in English. We randomly picked one such prompt to show this behavior.

\vspace{0.5em}
{\small
\begin{mdframed}[backgroundcolor=light-gray, roundcorner=2pt,leftmargin=0, rightmargin=0, innerleftmargin=5, innertopmargin=5,innerbottommargin=2, outerlinewidth=1, linecolor=light-gray]
\begin{lstlisting}
> Beskriv Eurovision.

Since its inception in 1956, the Eurovision Song Contest has become the largest music event in the world. The contest is a yearly event that brings together 40 countries to compete in a song contest. The winner is awarded the "Eurovision Song Contest Trophy".
The contest is hosted by a country that has won the contest in the previous year. The contest is broadcast live on television and radio to over 100 countries and is watched by over 100 million viewers. The contest is usually held in May or June.
The contest is a song contest, but there is a lot of cultural and political significance to it. The contest is a way for countries to showcase their music and culture. It is also a way for countries to show their support for one another.
The contest is a way for countries to show their support for one another. The contest is a way for countries to show their support for one another.
\end{lstlisting}
\end{mdframed} 
}

\paragraph{Training step 25 (3\,200 training samples)} The model starts to respond in Norwegian more confidently, even though the content is not always very helpful.

\vspace{0.5em}
{\small
\begin{mdframed}[backgroundcolor=light-gray, roundcorner=2pt,leftmargin=0, rightmargin=0, innerleftmargin=5, innertopmargin=5,innerbottommargin=2, outerlinewidth=1, linecolor=light-gray]
\begin{lstlisting}
> Beskriv Eurovision.

Hvert år arrangeres Eurovision Song Contest, en konkurranse mellom europeiske land. Hvert land velger en sang, og de beste sangene går videre til finalen. Det landet som vinner finalen, er det landet som får arrangere neste års konkurranse.
\end{lstlisting}
\end{mdframed} 
}
\vspace{-1em}
(\textit{Every year, the Eurovision Song Contest is held, a competition between European countries. Each country chooses a song, and the best songs go on to the final. The country that wins the final is the country that gets to host next year's competition.})

\paragraph{Training step 50 (6\,400 samples)} The responses are now always in Norwegian and their quality steadily improves.
\vspace{0.5em}
{\small
\begin{mdframed}[backgroundcolor=light-gray, roundcorner=2pt,leftmargin=0, rightmargin=0, innerleftmargin=5, innertopmargin=5,innerbottommargin=2, outerlinewidth=1, linecolor=light-gray]
\begin{lstlisting}
> Beskriv Eurovision.

Det er en årlig musikkonkurranse som arrangeres av Den europeiske kringkastingsunion. Målet er å kåre den beste sangen i Europa. Hvert land sender inn sitt bidrag, som består av en sang og en musikkvideo. Bidragene blir bedømt av en jury i hvert land, og av TV-seerne. Vinneren er den som får flest stemmer.
\end{lstlisting}
\end{mdframed} 
}
\vspace{-1em}
(\textit{It is an annual music competition organized by the European Broadcasting Union. The goal is to select the best song in Europe. Each country submits its entry, which consists of a song and a music video. The entries are judged by a jury in each country and by TV viewers. The winner is the one who receives the most votes}.)

\paragraph{Training step 75 (9\,600 samples)} The fluency scores have converged, the response quality is still improving.

\vspace{0.5em}
{\small
\begin{mdframed}[backgroundcolor=light-gray, roundcorner=2pt,leftmargin=0, rightmargin=0, innerleftmargin=5, innertopmargin=5,innerbottommargin=2, outerlinewidth=1, linecolor=light-gray]
\begin{lstlisting}
> Beskriv Eurovision.

Det europeiske musikkarrangementet Eurovision, også kjent som Eurovision Song Contest, er en årlig musikkonkurranse som arrangeres av Den europeiske kringkastingsunion (EBU). Konkurransen har blitt arrangert siden 1956, og er en av de lengstlevende musikkonkurransene i verden.

Eurovision er en direktesendt sangkonkurranse der hvert av de deltakende landene sender en sang som fremføres på scenen. Hvert land har en jury som stemmer på de andre landenes bidrag, og det landet som får flest stemmer, vinner konkurransen.

I tillegg til sangkonkurransen, er det også et stort show rundt arrangementet, med blant annet en åpningsseremoni, pressekonferanser og en finalefest. Arrangementet er svært populært i Europa, og tiltrekker seg millioner av seere hvert år.
\end{lstlisting}
\end{mdframed} 
}
\vspace{-1em}
(\textit{The European music event Eurovision, also known as the Eurovision Song Contest, is an annual music competition organized by the European Broadcasting Union (EBU). The competition has been held since 1956 [\dots]})

\paragraph{Training step 100 (12\,800 samples)} No apparent qualitative changes.

\vspace{0.5em}
{\small
\begin{mdframed}[backgroundcolor=light-gray, roundcorner=2pt,leftmargin=0, rightmargin=0, innerleftmargin=5, innertopmargin=5,innerbottommargin=2, outerlinewidth=1, linecolor=light-gray]
\begin{lstlisting}
> Beskriv Eurovision.

I Eurovision Song Contest, også kjent som Eurovision, samles land fra hele Europa for å konkurrere i en årlig musikkonkurranse. Konkurransen har blitt arrangert siden 1956 og består av to deler: en semifinale og en finale. I semifinalen fremfører hvert land en sang, og de ti beste går videre til finalen. I finalen stemmer hvert land på de andre landenes sanger, og vinneren er den som får flest stemmer.
Vinneren av Eurovision blir kåret gjennom en kombinasjon av telefonstemmer fra seerne og stemmer fra profesjonelle juryer. I tillegg til sangkonkurransen, er Eurovision også en plattform for å vise frem vertslandets kultur, og det arrangeres en rekke fester og arrangementer i forbindelse med konkurransen.
\end{lstlisting}
\end{mdframed} 
}
\vspace{-1em}
(\textit{In the Eurovision Song Contest, also known as Eurovision, countries from all over Europe come together to compete in an annual music competition. The contest has been held since 1956 and [\dots]}.)

\subsection{SFT outputs throughout training}
\label{app:samples-sft}

\paragraph{Training step 0} The SFT training stage starts from the same checkpoint as RL, so the responses (and their issues) are the same as in the previous section.

\vspace{0.5em}
{\small
\begin{mdframed}[backgroundcolor=light-gray, roundcorner=2pt,leftmargin=0, rightmargin=0, innerleftmargin=5, innertopmargin=5,innerbottommargin=2, outerlinewidth=1, linecolor=light-gray]
\begin{lstlisting}
> Beskriv Eurovision.

Since its inception in 1956, the Eurovision Song Contest has become the largest music event in the world. The contest is a yearly event that brings together 40 countries to compete in a song contest. The winner is awarded the "Eurovision Song Contest Trophy".
The contest is hosted by a country that has won the contest in the previous year. The contest is broadcast live on television and radio to over 100 countries and is watched by over 100 million viewers. The contest is usually held in May or June.
The contest is a song contest, but there is a lot of cultural and political significance to it. The contest is a way for countries to showcase their music and culture. It is also a way for countries to show their support for one another.
The contest is a way for countries to show their support for one another. The contest is a way for countries to show their support for one another.
\end{lstlisting}
\end{mdframed} 
}

\paragraph{Training epoch 1 (10\,000 training samples)} Similarly to on-policy RL, the model quickly starts to respond in Norwegian, according to the SFT training data. However, the texts often contain translationese artifacts. In this case, \textit{Basert på opprinnelsen} (based on [its] origin) sounds unnatural, this type of cataphoric thematization is more common in English. Similarly with \textit{sin storhet og prestisje} (its grandeur and prestige). Nominalizations should be generally used sparingly in favor of adjectives and verbs.

\vspace{0.5em}
{\small
\begin{mdframed}[backgroundcolor=light-gray, roundcorner=2pt,leftmargin=0, rightmargin=0, innerleftmargin=5, innertopmargin=5,innerbottommargin=2, outerlinewidth=1, linecolor=light-gray]
\begin{lstlisting}
> Beskriv Eurovision.

Basert på opprinnelsen er Eurovision Song Contest en årlig musikkonkurranse arrangert av Den europeiske kringkastingsunion (EBU) siden 1956. Den er kjent for sin storhet og prestisje, og tiltrekker seg oppmerksomhet fra hele verden. Konkurransen er en av de mest populære begivenhetene i Europa, og den er kjent for sin unike kombinasjon av musikk, show og underholdning.
\end{lstlisting}
\end{mdframed} 
}
\vspace{-1em}
(\textit{Based on its origin, the Eurovision Song Contest is an annual music competition organized by the European Broadcasting Union (EBU) since 1956. It is known for its grandeur and prestige, attracting attention from around the world. The contest is one of the most popular events in Europe and is known for its unique combination of music, showmanship, and entertainment})

\paragraph{Training epoch 2 (20\,000 training samples)}. The topicalization in the example below sounds very unnatural in Norwegian, almost ungrammatical. Using a relative clause would be more fluent in this case.

\vspace{0.5em}
{\small
\begin{mdframed}[backgroundcolor=light-gray, roundcorner=2pt,leftmargin=0, rightmargin=0, innerleftmargin=5, innertopmargin=5,innerbottommargin=2, outerlinewidth=1, linecolor=light-gray]
\begin{lstlisting}
> Beskriv Eurovision.

Først arrangert i 1956, er Eurovision Song Contest (ESC) en årlig musikkonkurranse arrangert av Den europeiske kringkastingsunion (EBU) blant medlemmene.
\end{lstlisting}
\end{mdframed} 
}
\vspace{-1em}
(\textit{First held in 1956, the Eurovision Song Contest (ESC) is an annual music competition organized by the European Broadcasting Union (EBU) among its members}.)

\paragraph{Training epoch 3 (30\,000 training samples)} The response below sounds natural.

\vspace{0.5em}
{\small
\begin{mdframed}[backgroundcolor=light-gray, roundcorner=2pt,leftmargin=0, rightmargin=0, innerleftmargin=5, innertopmargin=5,innerbottommargin=2, outerlinewidth=1, linecolor=light-gray]
\begin{lstlisting}
> Beskriv Eurovision.

Eurovision er en årlig musikkonkurranse arrangert av Den europeiske kringkastingsunion (EBU) med medlemmer fra over 50 land. Hvert land sender inn en sang, som deretter konkurrerer i en TV-sendt finale, som finner sted på forskjellige steder hvert år. Finalen består av to semi-finaler, hvor de ti beste fra hver semi-finale kvalifiserer seg til finalen. En jury og publikum bestemmer vinneren. Eurovision er et av de lengstlevende TV-programmene og tiltrekker seg rundt 200 millioner seere hvert år.
\end{lstlisting}
\end{mdframed} 
}
\vspace{-1em}
(\textit{Eurovision is an annual music competition organized by the European Broadcasting Union (EBU) with members from over 50 countries. Each country submits a song, which then competes in a televised final, held in different locations each year. The final consists of two semi-finals, with the top ten from each semi-final qualifying for the final. A jury and the audience decide the winner. Eurovision is one of the longest-running TV programs and attracts around 200 million viewers each year}.)

\paragraph{Training epoch 4 (40\,000 training samples)} In this case, the preposition in \textit{på en natt i mai} (on one night in May) is ungrammatical and likely a result of literal translation from English. Simply saying \textit{en natt i mai} would be correct.

\vspace{0.5em}
{\small
\begin{mdframed}[backgroundcolor=light-gray, roundcorner=2pt,leftmargin=0, rightmargin=0, innerleftmargin=5, innertopmargin=5,innerbottommargin=2, outerlinewidth=1, linecolor=light-gray]
\begin{lstlisting}
> Beskriv Eurovision.

Eurovision er en årlig musikkonkurranse arrangert av Den europeiske kringkastingsunion (EBU) der deltakerlandene konkurrerer med sanger på en natt i mai. Hvert land velger en sang som skal fremføres live foran et publikum og millioner av seere hjemme. Målet er å vinne konkurransen ved å få flest stemmer fra en jury og/eller seerne.
\end{lstlisting}
\end{mdframed} 
}
\vspace{-1em}
(\textit{Eurovision is an annual music competition organized by the European Broadcasting Union (EBU) in which participating countries compete with songs on one night in May. Each country selects a song to be performed live in front of an audience and millions of viewers at home. The goal is to win the competition by receiving the most votes from a jury and/or viewers}.)

\subsection{Outputs from the multilingual Mistral Nemo}
\label{app:samples-mistral}

\paragraph{Eurovision prompt} In this case, the multilingual model responded in Danish (which is correct since ``Beskriv Eurovision'' is ambiguous).

\vspace{0.5em}
{\small
\begin{mdframed}[backgroundcolor=light-gray, roundcorner=2pt,leftmargin=0, rightmargin=0, innerleftmargin=5, innertopmargin=5,innerbottommargin=2, outerlinewidth=1, linecolor=light-gray]
\begin{lstlisting}
> Beskriv Eurovision.

Eurovision Song Contest (ESC) er en årlig sangkonkurrence, der bliver afholdt af European Broadcasting Union (EBU) på vegne af de medlemslandes nationale tv-stationer. Konkurrencen blev første gang afholdt i 1956 og er dermed en af de ældste og mest sete tv-underholdningsprogrammer i Europa.

I Eurovision Song Contest konkurrerer to eller flere sanger fra hvert deltagende land om at blive kåret som den bedste sang. Sangen skal være skrevet specielt til konkurrencen og kan være op til tre minutter lang. Der er ingen begrænsninger i musikgenren, og sange kan være på ethvert sprog.

Vinderen af konkurrencen bestemmes af en jury af eksperter og af de stemmer fra publikum i de deltagende lande. Hvert land har en jury, der rangerer sangene fra 1 til 8, 10 og 12 point. Punktesystemet er baseret på det, der blev brugt i den første Eurovision Song Contest i 1956.

Eurovision Song Contest er kendt for sin glamour, underholdning og internationalitet. Konkurrencen er blevet beskrevet som et symbol på fred og forståelse mellem de europæiske lande og en chance for at vise og fejre det bedste i europæisk musik.
\end{lstlisting}
\end{mdframed} 
}
\vspace{-1em}
(\textit{The Eurovision Song Contest (ESC) is an annual song contest organized by the European Broadcasting Union (EBU) on behalf of the member countries' national television stations [\dots]})
\paragraph{Unambiguous prompt} To give a better example to the reader, we also show a response to a longer prompt that is unambiguously Norwegian, but containing several ungrammatical noun phrases, such as \textit{deres intime og personlige tonen}.

\vspace{0.5em}
{\small
\begin{mdframed}[backgroundcolor=light-gray, roundcorner=2pt,leftmargin=0, rightmargin=0, innerleftmargin=5, innertopmargin=5,innerbottommargin=2, outerlinewidth=1, linecolor=light-gray]
\begin{lstlisting}
> Hva regnes som det kjenteste verket i norsk litteratur fra poetisk realisme i perioden mellom 1850 og 1870?

Det mest kjente verket i norsk litteratur fra poetisk realisme i perioden mellom 1850 og 1870 er sannsynligvis «Fante-Breve» av Henrik Ibsen. «Fante-Breve» er en samling av brev som Ibsen skrev til sin kjæreste og senere hustru, Suzannah Thoresen, mens han var i Italia i 1864. Brevene er kjente for deres intime og personlige tonen, og de gir et innblikk i Ibsens tanker og følelser under denne perioden. «Fante-Breve» er også kjente for deres beskrivelser av den italienske kulturen og landskapet, og de har blitt sett på som et viktig verk innenfor norsk reise litteratur.
\end{lstlisting}
\end{mdframed} 
}
\vspace{-1em}
(\textit{What is considered the most famous work of Norwegian literature from the poetic realism period between 1850 and 1870? The most famous work of Norwegian literature from the period of poetic realism between 1850 and 1870 is probably “Fante-Breve” by Henrik Ibsen. “Fante-Breve” is a collection of letters that Ibsen wrote to his girlfriend and [\dots]})

\end{document}